\documentclass[journal]{IEEEtran}
\usepackage[utf8]{inputenc}
\usepackage{cite}

\usepackage{svg}
\usepackage{amsmath}
\usepackage{amssymb}
\usepackage{hyperref}
\usepackage{relsize}%
\usepackage{graphicx}
\usepackage{subfig}
\usepackage{xcolor}
\usepackage[utf8]{inputenc}
\usepackage[english]{babel}
\usepackage{amsthm}
\usepackage{xurl}

\usepackage{algorithmicx}
\usepackage[ruled,vlined, linesnumbered]{algorithm2e}
\usepackage{comment} 
\newcommand{\ignore}[1]{}
\usepackage{color} 
\usepackage{multirow}
\usepackage[T1]{fontenc}

\usepackage{algpseudocode}
\usepackage{bm}
\usepackage{enumitem}
\usepackage{physics}

\makeatletter
\newcommand*{\rom}[1]{\expandafter\@slowromancap\romannumeral #1@}
\newcommand*{\Scale}[2][4]{\scalebox{#1}{$#2$}}%

\newcommand{\removelatexerror}{\let\@latex@error\@gobble}
\let\oldnl\nl
\newcommand{\nonl}{\renewcommand{\nl}{\let\nl\oldnl}}
\makeatother

\IEEEoverridecommandlockouts
\begin{document}

\title{\LARGE \bf Design and Nonlinear Modeling of a Modular Cable Driven \\Soft Robotic Arm}

\author{Xinda Qi$^{1,*}$,
Yu Mei$^{1}$, Dong Chen$^{1}$ , Zhaojian Li$^{{2}}$,  \textit{Senior Member, IEEE},  Xiaobo Tan$^{1}$, \textit{Fellow, IEEE}
\vspace{-15pt}

\thanks{This work has been submitted to the IEEE for possible publication. Copyright may be transferred without notice, after which this version may no longer be accessible.}
\thanks{This work was supported by the US National Institutes of Health (Grant 1UF1NS115817-01) and the National Science Foundation (Grants ECCS 2024649, CNS 2237577).}
\thanks{$^1$Xinda Qi, Yu Mei, Dong Chen, and Xiaobo Tan are with the Department of Electrical and Computer Engineering, Michigan State University, Lansing, MI, 48824, USA. Email:{\tt\{qixinda, chendon9, xbtan\}@msu.edu}.}
\thanks{$^2$Zhaojian Li is with the Department of Mechanical Engineering, Michigan State University, Lansing, MI, 48824, USA. Email:{\tt\ lizhaoj1@msu.edu}.}
\thanks{$^*$ Corresponding author.}}

\maketitle

\begin{abstract}
We propose a novel multi-section cable-driven soft robotic arm inspired by octopus tentacles along with a new modeling approach. Each section of the modular manipulator is made of a soft tubing backbone, a soft silicon arm body, and two rigid endcaps, which connect adjacent sections and decouple the actuation cables of different sections. The soft robotic arm is made with casting after the rigid endcaps are 3D-printed, achieving low-cost and convenient fabrication. To capture the nonlinear effect of cables pushing into the soft silicon arm body, which results from the absence of intermediate rigid cable guides for higher compliance, an analytical static model is developed to capture the relationship between the bending curvature and the cable lengths. The proposed model shows superior prediction performance in experiments over that of a baseline model, especially under large bending conditions. Based on the nonlinear static model, a kinematic model of a multi-section arm is further developed and used to derive a motion planning algorithm. Experiments show that the proposed soft arm has high flexibility and a large workspace, and the tracking errors under the algorithm based on the proposed modeling approach are up to 52$\%$ smaller than those with the algorithm derived from the baseline model. The presented modeling approach is expected to be applicable to a broad range of soft cable-driven actuators and manipulators.

\end{abstract}

\IEEEpeerreviewmaketitle

\section{Introduction}

Soft robotic manipulators have been widely proposed and developed for
their various advantages, such as safe human-machine interactions,
robustness, and flexibility.\cite{r1, r2, r3, r4, r5} Compared with their
fully rigid counterparts, soft manipulators are able to utilize the
compliance and softness of their body structures to adapt to external
collisions and constraints and mitigate potential risks to humans, while
being able to accomplish traditional manipulation
tasks.\cite{r6,r7,r8} The advantages of soft manipulators make
them competitive candidates for applications involving the handling of
delicate and complex objects, as in fruit harvesting and medical
surgeries.\cite{r9,r10,r11}

Multiple structures and actuation methods have been developed for
building soft robotics arms to achieve compliant and efficient
deformation. For example, fluid-driven methods are widely used for soft
actuators, where fluid pressures inside actuator chambers are modulated
to generate elastic deformation.\cite{r12,r13,r14} Soft actuators
have also been constructed with other actuation mechanisms including
smart materials such as electroactive polymers.\cite{r15,r16,r17,r18,r19}
In particular, the cable-driven actuation method is popular thanks to
its simplicity and high force-to-weight ratio,\cite{r20,r21,r22}
where embedded eccentric cables driven by motors deliver torques to
achieve deformation of the soft body.

To control the deformation of the soft arm effectively, models that
capture the relationship between the actuation space input and the task
space output of the robotic arm have been developed. This modeling
process is generally complex and often dependent on robot designs and
actuation methods. Models for fluid-driven actuators are often built
based on static and dynamic analyses.\cite{r23,r24} For simple
cable-driven actuators, models have been built based on geometry
relationships.\cite{r25} Static models for tendon-driven
flexible manipulators have also been proposed to analyze the deformation
of the elastic tendons.\cite{r26,r27} In addition, models have
been reported for the coupling and decoupling cable system of
multi-section soft manipulators.\cite{r25,r28} Piecewise
constant curvature (PCC) models are widely utilized due to their
simplicity,\cite{r25} while other models, such as finite
element method (FEM) models and Cosserat rod models are proposed with
better accuracy but higher complexity.\cite{r29,r30,r31,r32} In
addition, piecewise constant strain (PCS) and geometric variable strain
models have been proposed that allow the incorporation of external forces
and more general settings.\cite{r33,r34,r35,r36}

Many biological structures and mechanisms have inspired the design of
robotic systems, and conversely, the development of robotic platforms
has provided bio-physical models for understanding biology and
biomechanics.\cite{r37,r38,r39,r40} In this study, inspired by the
longitudinal muscles in the anatomy of octopus tentacles, we propose a
decoupled modular cable-driven soft robotic arm fabricated with 3D
printing and casting. We further develop a novel analytical static model
that considers the prominent nonlinear effect of the cable pushing into
the soft body of the robotic arm. To our best knowledge, this is
the first effort in explicitly modeling the transverse deformation effect in soft cable-driven actuators, which extensively exists and might significantly influence the actuation length of the cable.

\begin{figure*}[!ht]
  \centering
\includegraphics[width=0.95\textwidth]{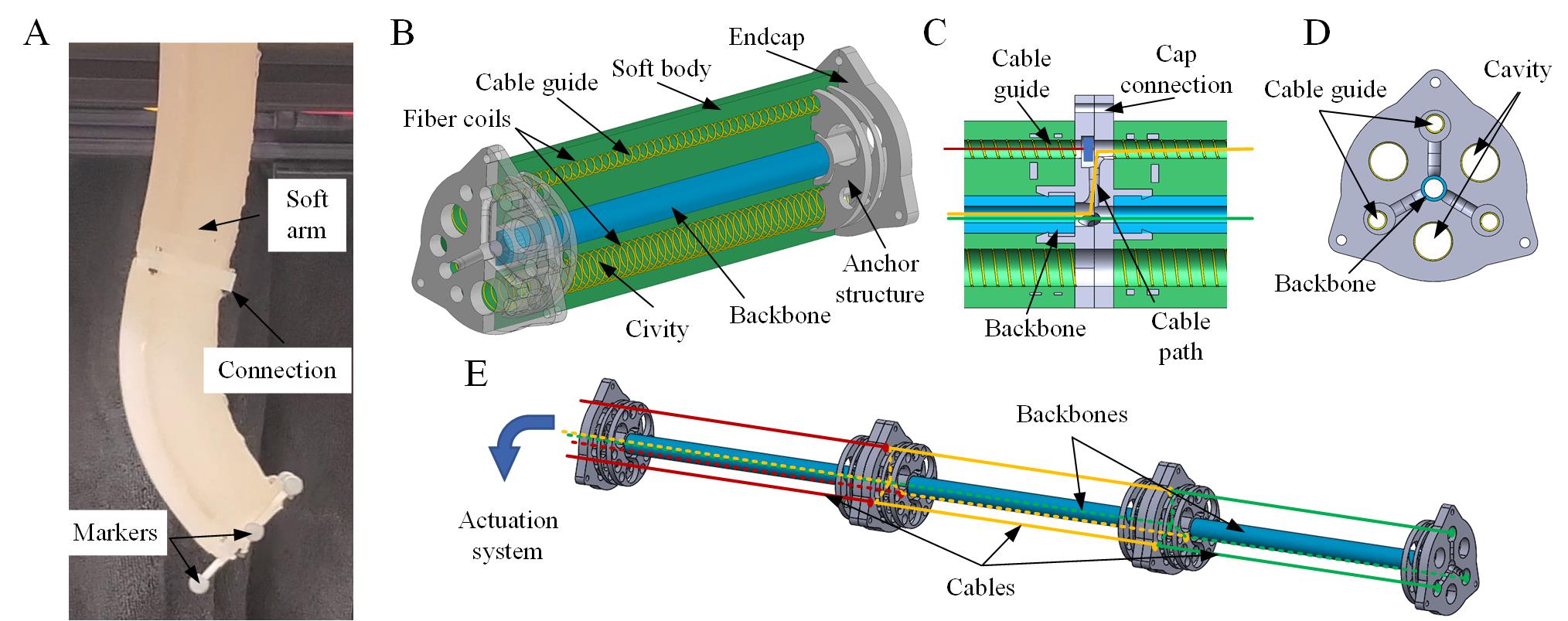}
  \caption{\small{\textbf{Structures of the soft robotic arm.} (A). A two-section modular soft robotic arm. (B). The structure for one section of the robotic arm. (C). Connection of two endcaps. (D) One endcap at the tip side of the section. (E). Cable paths for different sections of the soft robotic arm.}}
  \label{fig:figure}
  \vspace{-10pt}
\end{figure*}

Specifically, the structure of a single section of the proposed robotic
arm consists of three parts: a flexible backbone, a soft silicone body,
and two rigid caps (Fig. 1B). A piece of soft tubing is selected as the
backbone for its high bending flexibility and low stretchability to
constrain the section length. Two rigid endcaps are attached to the ends
of the backbone, which act as connectors between sections and anchor
points for cables. The anchor structure is designed to achieve a robust
connection between the endcaps and the soft body. The soft silicone body
is made by casting with three evenly embedded fiber-reinforced cable
guides and cavities. Actuation cables in the cable guides provide
contraction forces like longitudinal muscles while the cavities are able
to reduce the bending stiffness of the soft arm.

The soft modular multi-section robotic arm consists of identical
sections with the embedded cable system. The connection of the endcaps
is able to generate pathways between the cable guides and the backbone
tubing (Fig. 1C-D). The actuation cables, with their one end fixed on
the endcap, pass through the cable guides in one section and go into the
backbone tubing through the pathways before they are attached to the
corresponding driving motors (Fig. 1E). When one section of the arm has
a bending deformation, its backbone tubing maintains a nearly constant
length, ensuring the total lengths of the actuation cables for other
sections do not passively change. During the bending motion of the
multi-section robotic arm, the backbone tubing protects the actuation
system for other sections, separating the deformation of one section
from the change of cable lengths of other sections, and thus achieving
decoupling between different sections of the robotic arm.

\section{Materials and Methods}

The proposed soft robotic arm was modular and consisted of multiple sections. For each section, the fabrication process was separated into two steps: 3D printing and casting (Fig. 2). The rigid endcaps and the casting mold for the soft sections were first printed by a high-precision Object Connex 350 3D printer, which could construct complex structures layer by layer by jetting photopolymerizable materials that were cured by subsequent UV light. The material Objet Vero White was used in the 3D printing process for the rigid parts. The casting molds for the soft section including the rods for creating the fluid cavities were also fabricated by using the same 3D printing method. After all the molds and rigid parts for the soft section were prepared, the flexible backbone tubing (Clear Masterkleer Soft PVC Plastic Tubing, McMaster-Carr) was connected with the two rigid endcaps before assembling with the enclosure structures of the casting mold, which included a sole plate and three separate enclosure walls for easy removal of the molds after casting. Then, a high-strength Kevlar thread (High-Strength High-Temperature Thread, McMaster-Carr) was used to create a coil layer around the rods for the fluid cavities and the cable guides. The rods with the fiber coils were then inserted into the assembled casting mold with the help of locating holes on the two endcaps to complete the mold assembly.  

\begin{figure*}[!ht]
  \centering
\includegraphics[width=0.95\textwidth]{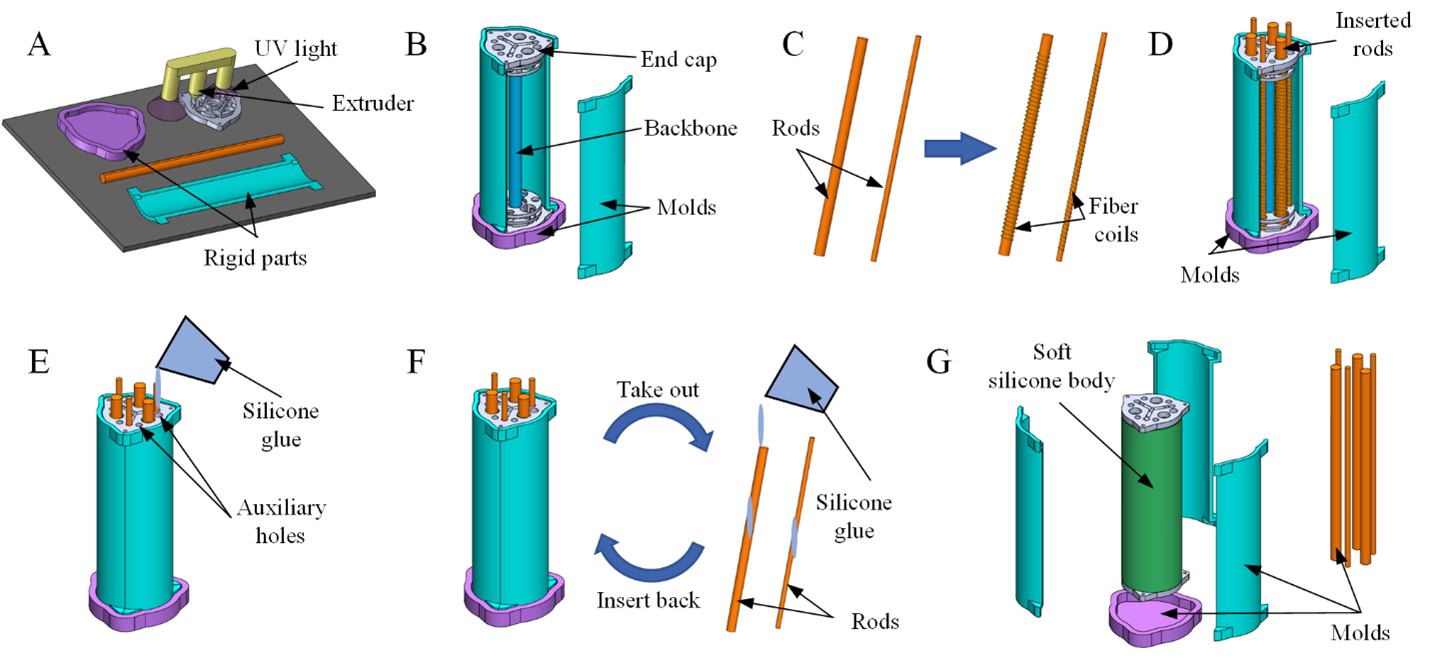}
  \caption{\small{\textbf{Fabrication process for one section of the soft robotic arm.} (A). 3D printing of the rigid parts. (B). Assembling of mold for one section. (C). Wrapping fiber coils around the rods for cavities. (D). Inserting the rods for cavities into the mold. (E). Injecting silicone glue into the mold and curing. (F). Adding another layer of silicone glue to the inner surface of the cavities. (G). Removing the mold and getting one section of the soft robotic arm.}}
  \label{fig:figure}
  \vspace{-10pt}
\end{figure*}

The silicone glue Ecoflex 00-10 was then used for the construction of the soft body of the arm section in the casting process and was injected into the mold from the axillary hole of one endcap. After the curing of the silicone glue, the rods for the cavities were extracted from the mold and were then inserted back with a silicone glue covering to construct an extra layer of silicone over the fiber coil for protection. Finally, all the mold pieces and the seal taps were removed and one section for the soft robotic arm was fabricated. After each section of the robotic arm was prepared, the actuation cables were first assembled with different sections. For each cable, one end was fixed on the endcap by using an anchor piece, and the other end was attached to the driving pulley. Screw connections were then applied to link different sections and the base for the whole robotic arm to complete the fabrication and assembling process of the prototype of the proposed soft robotic arm.

\section{Modeling of the Soft Robotic Arm}

The kinematic model for the multi-section soft robotic arm is separated
into two parts: a static model for a single section, which maps the
actuation cable lengths to the bending configuration of one section; and
another module that characterizes the relationship between the bending
configurations for all sections and the task space variables (in
particular, the end position of the arm).

\subsection{Static Model of a Section Driven by a Single Cable}

The static model for a single section of the robotic arm is built based
on the analysis of its bending deformation. Before studying a section
with multiple actuation cables, we consider the case with a single
actuating cable and analyze an arbitrary bending configuration (Fig.
3A). The cable and the support of the section provide external forces.
To simplify the static analysis, several assumptions are made:

\textbf{(A1)} The backbone (dash line) of the soft section has a constant length.
\textbf{(A2)} The backbone and the cable (red line) have constant curvatures.
\textbf{(A3)} The backbone has bending deformations within one single plane.
\textbf{(A4)} The friction between the cables and the soft body of the arm is
negligible.
\textbf{(A5)} The soft section of the arm has a linear bending stiffness with no
hysteresis.
\textbf{(A6)} The cables have no slacks.

Refer to Fig. 3A. Let \(s_{b}\) (subscript ``\emph{b}'' for
``backbone'') be the arclength parameter for the backbone. The bending
angle \(\phi\left( s_{b} \right)\) of the section at a given point with
an arclength \(s_{b}\) is defined as the angle of rotation between the
two local frames at the base,
\(\bm{e =}\left( \bm{e}_{\bm{x}}{,\ }\bm{e}_{\bm{y}} \right)\bm{,\ }\)and
at the point with \(s_{b}\) on the backbone,
\(\bm{a =}\left( \bm{a}_{\bm{x}}{,\ }\bm{a}_{\bm{y}} \right)\){,}
and can be described as: 
\begin{equation}
    \phi\left( s_{b} \right) = \frac{s_{b}}{R_{b}}
\end{equation}
where \(R_{b}\) is the radius of the backbone curvature. The curvature
of the backbone, \(\kappa_{b}\), is written as: 
\begin{equation}
    \kappa_{b} = \dv{\phi}{s_{b}} = \frac{1}{R_{b}}
\end{equation}

\begin{figure*}[!ht]
  \centering
\includegraphics[width=\textwidth]{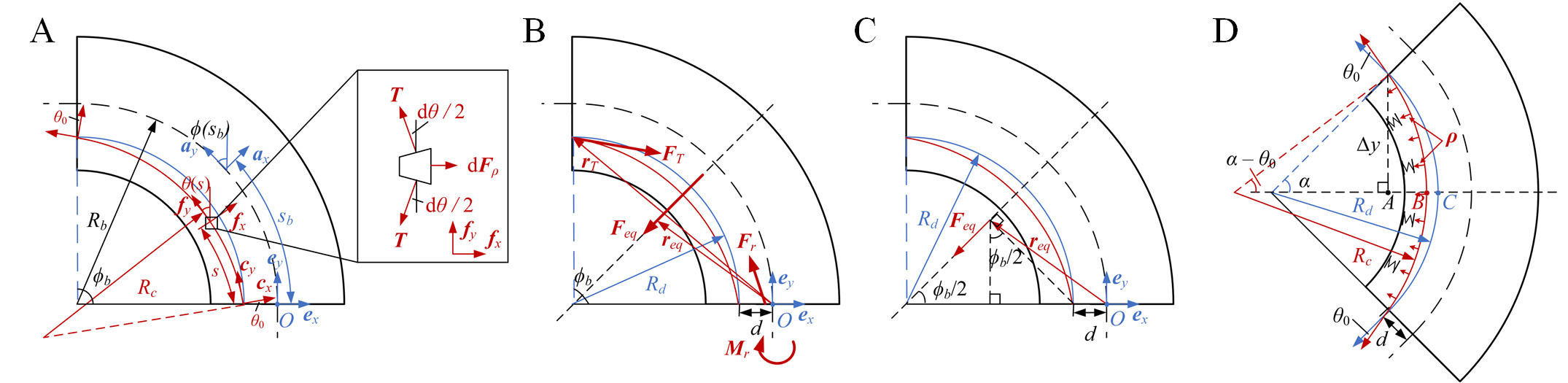}
  \caption{\small{\textbf{Modeling of one section of the arm driven by a single cable.} (A). Bending configuration for one section driven by a single cable. (B). External forces and moments applied by the cable to the soft section. (C) The total transverse force applied by the cable and its arm. (D). Actuation cable considering transverse deformation (red) and without transverse deformation (blue).}}
  \label{fig:figure}
  \vspace{-10pt}
\end{figure*}

The curvature \(\kappa_{c}\) (subscript ``\emph{c}'' for ``cable'') for
the cable is described similarly: \(\kappa_{c} = 1\text{/}R_{c}\),
where \(R_{c}\) is the radius of the actuation cable curvature, and
\(s\) is the arclength parameter for the cable. As illustrated in Fig.
3A, \(\theta(s)\) is the rotation angle between the base frame
\emph{\textbf{e}} and $\bm{f} = (\bm{f}_x, \bm{f}_y)$,
where \emph{\textbf{f}} is the local frame at the point with an
arclength of \(s\) on the cable.

Based on the force balance (Fig. 3A), the transverse force density
\(\rho\) between the cable and the soft body (Fig. 3D) is derived as:
\begin{equation}
    \rho = \dv{F_{\rho}}{s} = T\kappa_{c}
\end{equation}
\begin{equation}
    {\rm d}F_{\rho} = T{\rm d}\theta
\end{equation}

where \(T\) is the tension of the cable, \(F_{\rho}\) is the transverse
force between the cable and the soft body, \(\theta\) is used to denote
\(\theta(s)\) for simplicity, the notation ``d'' represents
differential, and the relationship \(\dv{\theta}{s} = \kappa_{c}\)
is used in the derivation of (3).

Another assumption following the introduction of \(\rho\) is made to
describe a simplified interaction model between the cable and the soft
body of the arm:

\textbf{(A7)} The maximum transverse deformation of the cable (\(|BC|\) in Fig.
3D) is proportional to the transverse force density \(\rho\) applied by
the cable (Fig. 3D).

Next, the transverse force density vector \(\bm{\rho}^{e}\) at point
\emph{s}, viewed in the base frame \emph{\textbf{e}} (Fig. 3A), is
calculated by using a rotation matrix \(R_{f}^{e}\):
\begin{equation}
\rho^e(\theta)=R_f^e\left[\begin{array}{c}
-\rho \\
0
\end{array}\right]
\end{equation}
\begin{equation}
    R_f^e(\theta)=\left[\begin{array}{cc}
\cos \theta & -\sin \theta \\
\sin \theta & \cos \theta
\end{array}\right]
\end{equation}

The total transverse force $\bm{F}_{eq}$ (subscript ``\emph{eq''}
for ``equivalent total force'') between the cable and the soft body can
then be obtained by the following integration:
\begin{equation}
\begin{aligned}
\boldsymbol{F}_{e q} & =\int_0^l \boldsymbol{\rho}^e(\theta(s)) \mathrm{d} s \\
& =\int_{\theta_0}^{\phi_b-\theta_0} \boldsymbol{\rho}^e(\theta) \frac{\mathrm{d} \theta}{\kappa_c} \\
& =T\left[\begin{array}{c}
-\sin \left(\phi_b-\theta_0\right)+\sin \theta_0 \\
\cos \left(\phi_b-\theta_0\right)-\cos \theta_0
\end{array}\right]
\end{aligned}
\end{equation}
where \(l\) is the cable length in the soft section, \(\phi_{b}\) is the
bending angle of the backbone at the tip, and \(\theta_{0}\) is the
incident angle of the cable, which is the angle between the tangent line
(\emph{\textbf{c}\textsubscript{y}} axis) of the cable at the base
surface and the normal (\emph{\textbf{e}\textsubscript{y}} axis) of the
base surface (Fig. 3A, D).

The contraction force \(\bm{F}_{T}\) applied by the cable tip to the
soft section (Fig. 3B) is calculated as:
\begin{equation}
    \boldsymbol{F}_T=R_f^e\left(\phi_b-\theta_0\right) \cdot\left[\begin{array}{c}
0 \\
-T
\end{array}\right]=T\left[\begin{array}{c}
\sin \left(\phi_b-\theta_0\right) \\
-\cos \left(\phi_b-\theta_0\right)
\end{array}\right]
\end{equation}

From the force balance equation: 
\begin{equation}
    \boldsymbol{F}_r+\boldsymbol{F}_{e q}+\boldsymbol{F}_T=0
\end{equation}
where \emph{\textbf{F}\textsubscript{r}} (subscript ``\emph{r}'' for
``reaction'') represents the support force applied by the base support
of the soft section to the soft section (Fig. 3B), one can derive:
\begin{equation}
    \Scale[1]{
        \begin{aligned}
    \bm{F}_r =-\bm{F}_{e q}-\bm{F}_T =T\left[\begin{array}{c}
    -\sin \theta_0 \\
    \cos \theta_0
    \end{array}\right]
    \end{aligned}
    }
\end{equation}

Next, the moment balance of the section is analyzed with respect to the
base point \emph{O} (Fig. 3B). The arm \(\bm{r}_{T}\) (Fig. 3B) for
\(\bm{F}_{T}\) is derived as:
\begin{equation}
\begin{aligned}
\boldsymbol{r}_T & =\left[\begin{array}{c}
-d-\left(R_d-R_d \cos \phi_b\right) \\
R_d \sin \phi_b
\end{array}\right] \\
& =\left[\begin{array}{c}
-d-2 R_d \sin ^2\left(\phi_b / 2\right) \\
2 R_d \sin \left(\phi_b / 2\right) \cos \left(\phi_b / 2\right)
\end{array}\right]
\end{aligned}
\end{equation}

where \(R_{d}\) is the radius of the cable curvature when the transverse
deformation of the cable is not considered, \(d\) is the distance
between the incident point of the cable and the base point \emph{O} of
the section.

Since \(\bm{F}_{eq}\) is located on the mirror-symmetric axis of the
bending section (see Fig. 3C), the exact point of action of
\(\bm{F}_{eq}\) is irrelevant in computing the resulting moment
around the point \emph{O}; in other words, the moment is the same
regardless of the point of action of \(\bm{F}_{eq}\). We have thus
chosen the point of action as illustrated in Fig. 3C, with the
associated arm vector represented as:
\begin{equation}
\begin{aligned}
r_{e q} & =\left[\begin{array}{c}
-d-R_d \sin \left(\phi_b / 2\right) \sin \left(\phi_b / 2\right) \\
R_d \sin \left(\phi_b / 2\right) \cos \left(\phi_b / 2\right)
\end{array}\right] \\
& =\left[\begin{array}{c}
-d-R_d \sin ^2\left(\phi_b / 2\right) \\
R_d \sin \left(\phi_b / 2\right) \cos \left(\phi_b / 2\right)
\end{array}\right]
\end{aligned}
\end{equation}

From the moment the balance of the section: 
\begin{equation}
    \bm{M}_r+\bm{M}_{e q}+\bm{M}_T=0
\end{equation}
where \emph{\textbf{M}\textsubscript{r}} denotes the support moment,
\(\bm{M}_{eq}\) and \(\bm{M}_{T}\) are the moments generated by
\(\bm{F}_{eq}\) and \(\bm{F}_{T}\) with respect to point
\emph{O}, respectively, and:
\begin{equation}
    \boldsymbol{M}_{e q}+\boldsymbol{M}_T=\boldsymbol{r}_{e q} \times \boldsymbol{F}_{e q}+\boldsymbol{r}_T \times \boldsymbol{F}_T=T d \cos \theta_0
\end{equation}

one obtains: 
\begin{equation}
    \bm{M}_{r} = - Td\cos\theta_{0}
\end{equation}

See the Appendix for calculation details of this concise result. Next, the incident angle of the cable \(\theta_{0}\) (Fig. 3D) satisfies:
\begin{equation}
    \Delta y=\left(R_b-d\right) \sin \alpha=R_c \sin \left(\alpha-\theta_0\right)
\end{equation}
which implies:
\begin{equation}
    \theta_0=\alpha-\arcsin \left(\left(1-\kappa_b d\right) \frac{\kappa_c}{\kappa_b} \sin \alpha\right)
\end{equation}
where \(\alpha = \phi_{b}/2\) .

Based on assumption (A7) and the geometric relationships (Fig. 3D), one
can obtain:
\begin{equation}
    \Delta h=|B C|=R_d(1-\cos \alpha)-R_c\left(1-\cos \left(\alpha-\theta_0\right)\right)
\end{equation}
\begin{equation}
    \rho = K_{c}\Delta h
\end{equation}
where \(\Delta h\) is the maximum transverse deformation of the
cable and \(K_{c}\) is the coefficient in the simplified linear
relationship between \(\mathrm{\Delta}h\) and \(\rho\).The physical interpretation of $K_c$ is the “cutting-in stiffness”, where the interaction between the normal force of the cable and the soft body is treated as a linear spring. Admittedly, the assumed linear relationship is a simplification, as the relationship could be nonlinear especially if the deformation is large. Despite the limitation, the experiment results later in the paper (see Fig. 8B) show that this model for the interaction between the cable and the soft body is adequately accurate to describe the transverse deformation phenomenon.

Based on assumption (A5), the relationship between the bending
deformation and the external torque is derived by introducing a bending
stiffness \(K_{b}\):
\begin{equation}
    \left|\boldsymbol{M}_{e q}+\boldsymbol{M}_T\right|=\left|\boldsymbol{M}_r\right|=M_r=K_b \kappa_b
\end{equation}
where \(K_{b}\kappa_{b}\) is the internal elastic moment.

Finally, by using the equations (3), (15), (17-20), and the geometric
relationships, the model for a single soft section driven by a single
cable is captured by:
\begin{equation}
        \left\{\begin{array}{c}
    K_b \kappa_b=T d \cos \theta_0 \\
    \theta_0=\alpha-\arcsin \left(\left(1-\kappa_b d\right) \frac{\kappa_c}{\kappa_b} \sin \alpha\right) \\
    T=\frac{K_c}{\kappa_c}\left\{\left(\frac{1}{\kappa_b}-d\right)(1-\cos \alpha)-\frac{1}{\kappa_c}\left(1-\cos \left(\alpha-\theta_0\right)\right)\right\} \\
    l=R_c\left(\phi_b-2 \theta_0\right)=\frac{1}{\kappa_c}\left(L \kappa_b-2 \theta_0\right)
    \end{array}\right.
\end{equation}

where \(L\) is the length of the soft section's backbone. In the forward
mapping from actuation to the robotic arm configuration, the backbone
curvature \(\kappa_{b}\) is solved based on the cable length \(l\) by
using a nonlinear equation set solver and numerical methods (e.g.
``fsolve'' in MATLAB), while in the inverse problem \emph{l} is
calculated based on a desired reference \(\kappa_{b}\) by using the
nonlinear equation set solver. The incident angle \(\theta_{0}\), cable
curvature \(\kappa_{c}\) and cable tension \(T\) are intermediate
variables, while \(d\), \(L\), \(K_{c}\), and \(K_{b}\) are constants
and \(\phi_{b}\), \(\alpha\) are dependent on \(\kappa_{b}\). The
initial guess for the solution to the nonlinear equations is derived by
solving the last three equations in (21) when we assume that there is no
transverse deformation of the cables (blue curve in Fig. 3A):
\(\theta_{0} = 0\).

\subsection{Static Model of a Section Driven by Multiple Cables}
\begin{figure*}[!ht]
  \centering
\includegraphics[width=\textwidth]{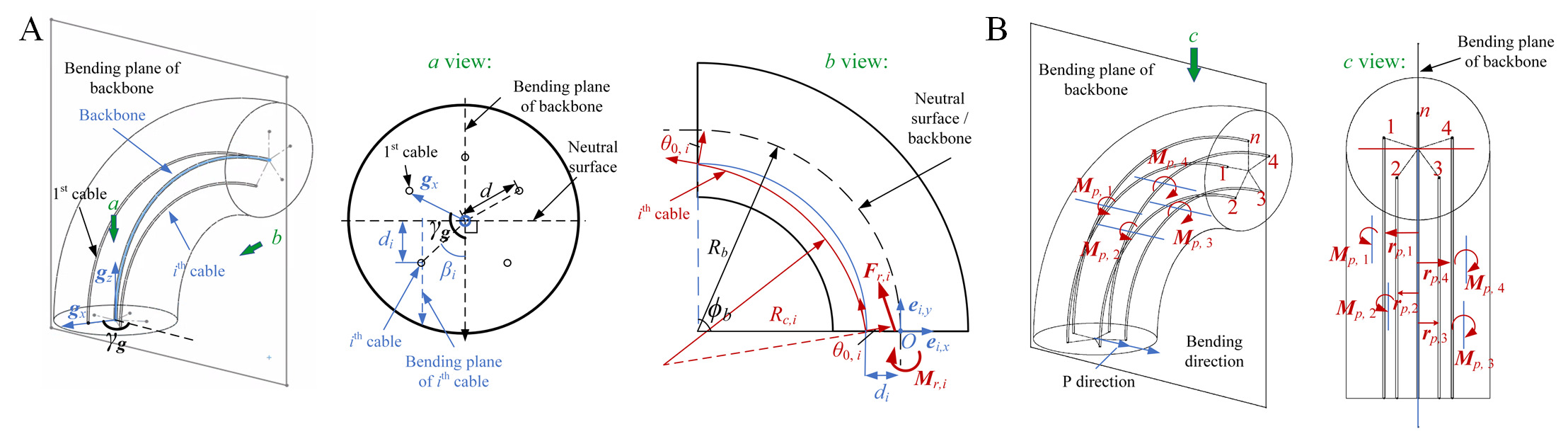}
  \caption{\small{\textbf{Modeling of one section of the arm driven by multiple cables.} (A). Bending configuration for one section of the arm driven by multiple cables. (B). External moments applied by multiple cables to the soft section in the $P$ direction.}}
  \label{fig:figure}
  \vspace{-10pt}
\end{figure*}

After the model for one section driven by one cable is obtained, the
model for the case of a single section with multiple actuating cables is
addressed, where we assume that there is no slack for any cable.

The bending configuration for the soft section with \emph{n} evenly
distributed cables (a general case) is defined by the bending angle
\(\phi_{b}\) and the bending orientation \(\gamma_{\bm{g}}\)
(\(\gamma_{\bm{g}}\) is with respect to the base frame
\emph{\textbf{g}}) (Fig. 4A). A curved neutral surface is defined so
that it is perpendicular to the bending plane and contains the backbone.
The cables are indexed counterclockwise and the direction of the
\emph{x} axis of the base frame \emph{\textbf{g}} points to the
1\textsuperscript{st} cable. The angle between the
\emph{i}\textsuperscript{th} cable orientation (in the base plane) and
the bending direction is written as (Fig. 4A):
\begin{equation}
    \beta_i=\frac{2 \pi(i-1)}{n}-\gamma_g \quad i=1,2,3 \ldots
\end{equation}
where \emph{n} is the total number of the evenly distributed cables in
the soft section.

The distance between the incident point of the
\emph{i}\textsuperscript{th} cable and the neutral plane is calculated
as:
\begin{equation}
    d_i=d \cos \beta_i \quad i=1,2,3 \ldots
\end{equation}

In the bending plane of the \emph{i}\textsuperscript{th} cable (Fig.
4A), which is parallel to the bending plane of the backbone and contains
the \emph{i}\textsuperscript{th} cable, the external force condition is
analyzed as the single cable-driven case:
\begin{equation}
    \Scale[0.95]{
        \left\{\begin{array}{c}
    \boldsymbol{F}_{r, i}=T_i\left[\begin{array}{c}
    -\sin \theta_{0, i} \\
    \cos \theta_{0, i}
    \end{array}\right] \\
    \left|\boldsymbol{M}_{r, i}\right|=\left|-T_i d_i \cos \theta_{0, i}\right| \\
    \theta_{0, i}=\alpha-\arcsin \left(\left(1-\kappa_b d_i\right) \frac{\kappa_{c, i}}{\kappa_b} \sin \alpha\right) \\
    T_i=\frac{K_c}{\kappa_{c, i}}\left\{\left(\frac{1}{\kappa_b}-d_i\right)(1-\cos \alpha)-\frac{1}{\kappa_{c, i}}\left(1-\cos \left(\alpha-\theta_{0, i}\right)\right)\right\}
    \end{array}\right.
    }
\end{equation}
where\(\ T_{i}\), \(\theta_{0,i}\), and \(\kappa_{c,i}\) are the cable
tension, incident angle and curvature of the
\emph{i}\textsuperscript{th} cable, respectively, and
\(\bm{F}_{r,i}\) and \(\bm{M}_{r,i}\) are the support force and
moment components induced by the \emph{i}\textsuperscript{th} cable,
respectively, in local frame \(\bm{e}_{i}\) (Fig. 4A).

Then, one can calculate the total bending moment applied by the cables
for the section:
\begin{equation}
\boldsymbol{M}=\sum_{i=1}^n-\boldsymbol{M}_{r, i}
\end{equation}
with:
\begin{equation}
|\boldsymbol{M}|=M=K_b \kappa_b
\end{equation}

Furthermore, since there is no bending deformation in the direction
perpendicular to the bending direction, and based on assumption (A3), the
total external moment applied by cables in the \emph{P} direction with
respect to \emph{O} (Fig. 4B) is zero:
\begin{equation}
    \sum_{i=1}^n \boldsymbol{M}_{p, i}=0
\end{equation}
where \(\bm{M}_{p,i}\) is the external moment applied by the
\emph{i}\textsuperscript{th} cable with respect to \emph{O}, in the
\emph{P} direction.

The arm of the external forces for \(\bm{M}_{p,i}\) is the distance
between the bending plane of the \emph{i}\textsuperscript{th} cable and
the bending plane of the backbone (Fig. 4B), whose direction is
perpendicular to the bending plane with the magnitude:
\begin{equation}
    r_{p, i}=d \sin \beta_i
\end{equation}

Thus, the lateral moment \(\bm{M}_{p,i}\) applied by the
\emph{i}\textsuperscript{th} cable is derived as:
\begin{equation}
    \boldsymbol{M}_{p, i}=\boldsymbol{F}_{T, i, y} \times \boldsymbol{r}_{p, i}+\int_0^{l_i} \boldsymbol{\rho}_{i, y}^e(s) \times \boldsymbol{r}_{p, i} d s=-\boldsymbol{F}_{r, i, y} \times \boldsymbol{r}_{p, i}
\end{equation}
where \(\bm{F}_{T,i,y}\), \(\bm{\rho}_{i,y}^{e}\), and
\(\bm{F}_{r,i,y}\) are the components along \(\bm{e}_{i,y}\)
axis (or \emph{\textbf{g\textsubscript{z}}} axis) of the
\(\bm{F}_{T,i}\), \(\bm{\rho}_{i}^{e}\), and
\(\bm{F}_{r,i}\), respectively,

Then, based on equations (23)-(29) and the geometric relationships, the
kinematic model for a soft section of the arm actuated by multiple
cables is derived as:
\begin{equation}
\Scale[0.95]{
\left\{\begin{array}{c}
\sum_{i=1}^n T_i d \cos \theta_{0, i} \cos \beta_i=K_b \kappa_b \\
\sum_{i=1}^n T_i d \cos \theta_{0, i} \sin \beta_i=0 \\
\theta_{0, i}=\alpha-\arcsin \left(\left(1-\kappa_b d_i\right) \frac{\kappa_{c, i}}{\kappa_b} \sin \alpha\right) \\
T_i=\frac{K_c}{\kappa_{c, i}}\left\{\left(\frac{1}{\kappa_b}-d_i\right)(1-\cos \alpha)-\frac{1}{\kappa_{c, i}}\left(1-\cos \left(\alpha-\theta_{0, i}\right)\right)\right\} \\
l_i=R_{c, i}\left(\phi_b-2 \theta_{0, i}\right)=\frac{1}{\kappa_{c, i}}\left(L \kappa_b-2 \theta_{0, i}\right)
\end{array}\right.
}
\end{equation}

In the forward mapping from actuation to arm configuration, given the
cable lengths \(l_{i}\), the backbone curvature \(\kappa_{b}\) and the
bending orientation \(\gamma_{\bm{g}}\) (manifested via
\(\beta_{i}\) in equation (22)) are obtained by solving the equation
(30). \(\theta_{0,i}\), \(\kappa_{c,i}\), and \(T_{i}\) are intermediate
variables, \(d\), \(L\), \(K_{c}\), and \(K_{b}\) are constants, and
\(\phi_{b}\), \(\alpha\) are fully determined by \(\kappa_{b}\).
Therefore, in the forward mapping problem, there are \((3n + 2)\)
unknowns and \((3n + 2)\) independent equations. In the inverse mapping
problem, the lengths \(l_{i}\) for different cables are calculated based
on the desired reference values of \(\kappa_{b}\) and
\(\gamma_{\bm{g}}\) by solving the same equation (30). It is noticed
that \(n\) (the number of the cables) is required to be larger than 2 to
achieve a redundant actuation system for all of the bending orientations
considering the limitation that \(T_{i}\) is non-negative (in the
inverse mapping, considering the equation (30), substituting
\(\beta_{i}\) by using \(\gamma_{\bm{g}}\), the numbers of
independent equations and the unknown variables are \((3n + 2)\) and
\((4n)\), respectively). The initial guess of the nonlinear system is
obtained by solving the last three rows in (30), where we assume
that there is no transverse deformation of the cables:
\(\theta_{0,i} = 0\).

\subsection{Modeling of a Multi-Section Soft Arm}

For the multi-section soft robotic arm, the kinematic model between the
task space configuration (in particular, the end position) and the
bending configuration for each section is built by using homogeneous
transformation matrices.\textsuperscript{25} Specifically, considering
the thickness of the rigid endcaps, one can divide each section of the
arm into three parts: straight, bending, and straight. The
transformation matrix \(T_{i,e}^{i,s}\), which transforms vectors in the
end frame \(\Sigma_{i,e}\) to those in the base frame \(\Sigma_{i,s}\)
for the \emph{i}\textsuperscript{th} section (Fig. 5A), is given by:
\begin{equation}
    \Scale[0.82]{
    T_{i, e}^{i, s}=
\begin{bmatrix}
I & \boldsymbol{p}_{0, i} \\
0 & 1
\end{bmatrix}
\begin{bmatrix}
R_z\left(\gamma_i\right) & 0 \\
0 & 1
\end{bmatrix}
\begin{bmatrix}
R_y\left(\phi_{b, i}\right) & \boldsymbol{p}_i \\
0 & 1
\end{bmatrix}
\begin{bmatrix}
R_z\left(-\gamma_i\right) & 0 \\
0 & 1
\end{bmatrix}
\begin{bmatrix}
I & \boldsymbol{p}_{0, i} \\
0 & 1
\end{bmatrix}
}
\end{equation}
where \(\phi_{b,i}\) and \(\gamma_{i}\) (\(\gamma_{i}\) is with respect
to a general base frame \(\Sigma_{i,s}\) that is not dependent on the
cable distribution) are the bending angle and orientation for the
\emph{i}\textsuperscript{th} section, respectively,
\(\bm{p}_{i} = \ \begin{bmatrix}
(1 - \cos{\phi_{b,i})/\kappa_{b,i}} & 0 & \sin{\phi_{b,i}/\kappa_{b,i}} \\
\end{bmatrix}^{T}\) is the in-plane displacement from the base to tip
for the \emph{i}\textsuperscript{th} section, \(\kappa_{b,i}\) is the
curvature of the backbone of the \emph{i}\textsuperscript{th} section,
\(\bm{p}_{0,i} = \ \begin{bmatrix}
0 & 0 & h \\
\end{bmatrix}^{T}\) is the displacement of the straight part and \(h\)
is the thickness of the endcaps. \emph{I} is a three-dimensional
identity matrix, \(R_{z}(\gamma_{i})\), \(R_{y}(\phi_{b,i})\), and
\(R_{z}( - \gamma_{i})\) are the 3D rotation matrices around the z-axis,
y-axis and z-axis for the angle \(\gamma_{i}\), \(\phi_{b,i}\), and
\(- \gamma_{i}\), respectively.

\begin{figure*}[!ht]
  \centering
\includegraphics[width=0.95\textwidth]{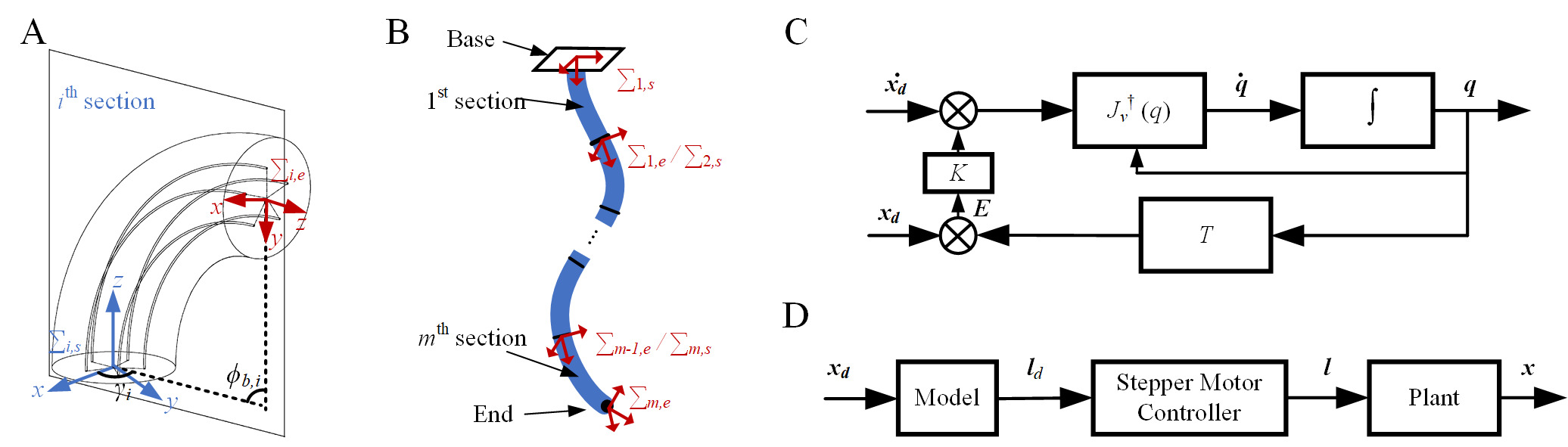}
  \caption{\small{\textbf{Modeling of a multi-section soft robotic arm.} (A). Variables of bending configuration for one section. (B). Local frames for different sections. (C). The inverse kinematics solver for the bending configurations of different sections based on the reference of the end position. (D). Open loop control of the soft robotic arm based on the proposed model. $\bm{x}$ and $\bm{l}$ are the end position of the soft robotic arm and the actuation cable lengths, respectively, for which $\bm{x}_d$ and $\bm{l}_d$ are the corresponding target values.}}
  \label{fig:figure}
  \vspace{-10pt}
\end{figure*}

The transformation matrix \(T_{m,e}^{1,s}\) which transforms vectors in
the end frame \(\Sigma_{m,e}\) to those in the base frame
\(\Sigma_{1,s}\) of the \emph{m}-section arm, and the end position of
the arm \(\bm{p}_{t}^{1,s}\) (subscript \emph{``t''} for ``tip'') in
\(\Sigma_{1,s}\) (Fig. 5B) can be calculated as:
\begin{equation}
    T_{m, e}^{1, s} =\prod_{i=1}^m T_{i, e}^{i, s}
\end{equation}
\begin{equation}
    \boldsymbol{p}_t^{1, s} =T_{m, e}^{1, s} \cdot \boldsymbol{p}_t^{m, e}
\end{equation}
where $\bm{p}_t^{m, e}=\begin{bmatrix}
   0 & 0 & 0 & 1 
\end{bmatrix}^T $.

For the inverse kinematics,\cite{r41} the linear velocity
Jacobian matrix \(J_{v}\) for the end position of the \emph{m}-section
arm with respect to the variables of the bending configuration of each
section is calculated as:
\begin{equation}
        J_v=J_{\boldsymbol{p}_t^{1, s}}(\boldsymbol{q})=\left[\begin{array}{lll}
    \frac{\partial \boldsymbol{p}_t^{1, s}}{\partial q_1} & \ldots & \frac{\partial \boldsymbol{p}_t^{1, s}}{\partial q_{2 m}}
    \end{array}\right]
\end{equation}
where $\boldsymbol{q}=\begin{bmatrix}
\gamma_1 & \kappa_{b, 1} & \ldots & \gamma_m & \kappa_{b, m}
\end{bmatrix}^T$ is the set of variables of the bending
configurations of all sections. The calculated \(J_{v}\) is omitted for
brevity and a small value \(\delta_{\kappa}\) is added to
\(\kappa_{b,i}\) when \(\kappa_{b,i} \rightarrow 0\) for numerical
stability. At least two sections are required for the arm to provide
redundancy for tracking desired end positions in 3D space.

Once \(J_{v}\) is obtained, one can use the following method to approach
the configurations given the desired task space output (inverse
kinematics) by using the pseudo-inverse of the Jacobian matrix
\({J_{v}}^{\dagger}\):
\begin{equation}
    \dot{\boldsymbol{q}}=J_v{ }^{\dagger} \mathcal{V}+\left(I-J_v{ }^{\dagger} J_v\right) \dot{\boldsymbol{q}}_0
\end{equation}
\begin{equation}
    J_v{ }^{\dagger}=J_v{ }^T\left(J_v J_v{ }^T+k^2 I\right)^{-1}
\end{equation}
where \(I\) is an identity matrix, \(k\) is a small positive number,
\(\mathcal{V} = {\dot{\bm{x}}}_{d}(t)\) is the velocity vector of
the tracking trajectory \(\bm{x}_{d}(t)\) (subscript \emph{``d''}
for ``desired''), \({\dot{\bm{q}}}_{0}\) is any vector with the
shape of \(\dot{\bm{q}}\) and set to zero for the minimum energy
criterium.

By using a numerical method to integrate the velocities, the references
for the bending configuration variables are calculated:
\begin{equation}
    \boldsymbol{q}\left(t_{k+1}\right)=\boldsymbol{q}\left(t_k\right)+J_v{ }^{\dagger}\left(\boldsymbol{q}\left(t_k\right)\right) \cdot \dot{\boldsymbol{x}}_d\left(t_k\right) \cdot \Delta t
\end{equation}
where \(\bm{q}\left( t_{k} \right)\) and
\(\bm{q}\left( t_{k + 1} \right)\) denote \(\bm{q}\) at the time
steps \(t_{k}\) and \(t_{k + 1}\), respectively.

Closed-loop feedback is further implemented (Fig. 5C) to reduce the
error accumulated by the numerical integration in (37):
\begin{equation}
    \dot{\boldsymbol{q}}=J_v{ }^{\dagger}(\boldsymbol{q})\left(\dot{\boldsymbol{x}}_d+K \cdot \boldsymbol{E}\right)
\end{equation}
where
\(\bm{E} = \bm{x}_{d} - T_{m,e}^{1,s}(\bm{q}) \cdot \bm{p}_{t}^{m,e}\)
is the feedback error and \(K\) is a positive diagonal gain matrix. Once
\(\bm{q}\) is obtained, we can use the equation (30) to calculate
the cable lengths \(l_{i}\) for different actuation cables for all
sections.

\section{Results}

\subsection{Baseline Model for the Soft Robotic Arm}

Extensive experiments have been conducted to validate the proposed
model. For comparison, an existing baseline model\textsuperscript{25}
for the soft robotic arm is built, without considering the transverse
deformation of the cable during bending. The multi-section part of the
baseline kinematic model is the same as that of the proposed model, but
the relationship between the cable lengths and the bending configuration
of a single section is derived based on ideal geometrical relationships
for the baseline model. In particular, the curvature \(\kappa_{c,i}\)
and the length \(l_{i}\) of the \emph{i}\textsuperscript{th} cable in a
section (blue curve in Fig. 4A) is derived as:
\begin{equation}
    \frac{1}{\kappa_b}=\frac{1}{\kappa_{c, i}}+d_i
\end{equation}
\begin{equation}
    l_i=R_{c, i} \cdot \phi_b=\frac{\kappa_b}{\kappa_{c, i}} L
\end{equation}
where \(d_{i}\) is derived from equation (23), \(\kappa_{b}\) and \(L\)
are the curvature and the length of the backbone, respectively.

\subsection{Experiment Setups}
During the experiments for the soft robotic arm, 5 optical markers were attached at the end of the robotic arm to monitor the bending configuration and the end position of the robotic arm with a motion capture system (Fig. 6). The motion capture system used in the experiments was ``Opti-track'', including a set of infrared cameras and the related software, which could track the position and orientation of the “Rigid Body” consisting of the attached markers on the robotic arm and provide accurate information for the bending angle $\phi_b$, bending orientation $\gamma$, and the position for the end of the robotic arm. 
\begin{figure}[!ht]
  \centering
\includegraphics[width=0.4\textwidth]{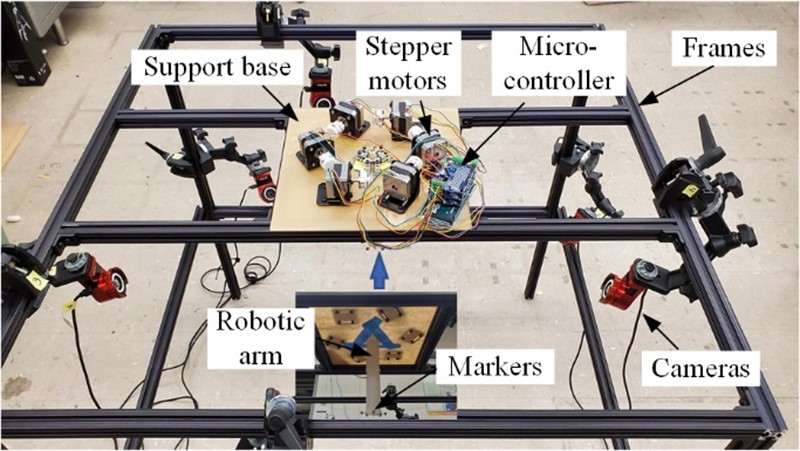}
  \caption{\small{Overall configuration of the experimental setups for the soft robotic arm.}}
  \label{fig:figure}
  \vspace{-10pt}
\end{figure}

In the experiments, the cable length of the cables was controlled by using stepper motors, and the bending angle $\phi_b$ and orientation $\gamma$ were recorded by the motion capture system. The actuation cable was driven by a 3D-printed pulley which was attached to a stepper motor (NEMA-17, Adafruit) (Fig. 7A). The stepper motors were controlled by a micro-controller (Arduino Mega 2560) in an open loop control (without encoder feedback) with the help of a stepper motor driver (Adafruit Motor Shield V2) for power amplification (Fig. 7B). Then, via serial communication, the microcontroller was communicated with a main controller (laptop), where the analytical models were computed before the reference for the stepper motor was sent out. 
\begin{figure}[!ht]
  \centering
\includegraphics[width=0.35\textwidth]{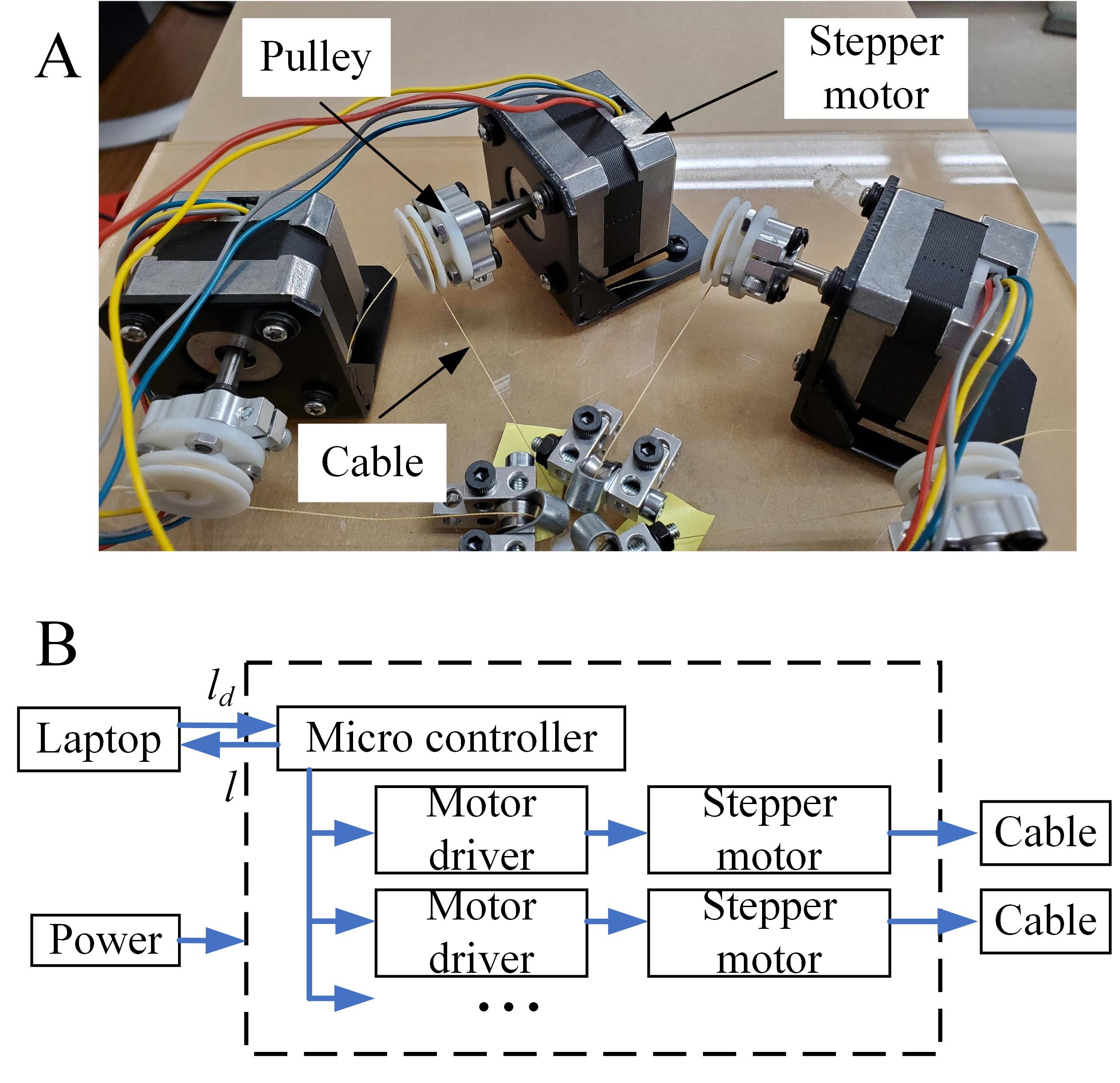}
  \caption{\small{Cable actuation mechanism for the soft robotic arm. (B). Diagram of the control and actuation system for the soft robotic arm.}}
  \label{fig:figure}
  \vspace{-10pt}
\end{figure}

\subsection{Parameter Identification}

The geometric parameters ($L, d$) were measured directly from the
prototype. The bending stiffness \(K_{b}\) for a single section of the
arm was calculated by using equations (15) and (20) assuming
\(\theta_{0} \approx 0\) when \(\phi_{b}\) was small, where \(T\) and
\(\phi_{b}\) were measured by a force sensor and a motion capture
system, respectively.

The relationship between a single cable contraction length \(\Delta l\)
and \(\phi_{b}\) was obtained in the experiments for one soft section
driven by a single cable to identify \(K_{c}\) by using equation (21). 
In the experiments, the cable length $l$ of a single cable was controlled by using one stepper motor, and the bending angle $\phi_b$ was recorded by the motion capture system. The speeds of the stepper motors were controlled to be relatively slow for a quasi-static condition in all experiments. The contraction cable length $\Delta l = L - l$ were calculated after the experiments, and the experimental relationship between the $\Delta l$ and $\phi_b$ was obtained (Fig. 8B), based on which $K_c$ was identified.

The experiment results and the estimations based on the baseline
model and the proposed model (equation (21)) with the identified
parameters are shown in Fig. 8B. The error bars in experimental results represent the standard deviation of the result in three runs for each bending configuration. The identified parameters of the one section for the soft robotic arm used in the experiments are shown in the Table 1.
\begin{table}[!ht]
    \caption{\small{The identified parameters of one section of the soft arm prototype used in the experiments}}
    \centering
    \begin{tabular}{c c c c c}
    \hline\hline \vspace{3pt}
       Parameters  & $L(cm)$ & $d (cm)$ & $K_b (N\cdot cm^2)$ & $K_c (N / cm^2)$\\
       \hline \vspace{2pt}
       Values  & 9.30 & 1.25 & 20.02 & 3.10 \\
    \hline\hline
    \end{tabular}
    \label{tab:my_label}
\end{table}

\begin{figure*}[!ht]
  \centering
\includegraphics[width=0.9\textwidth]{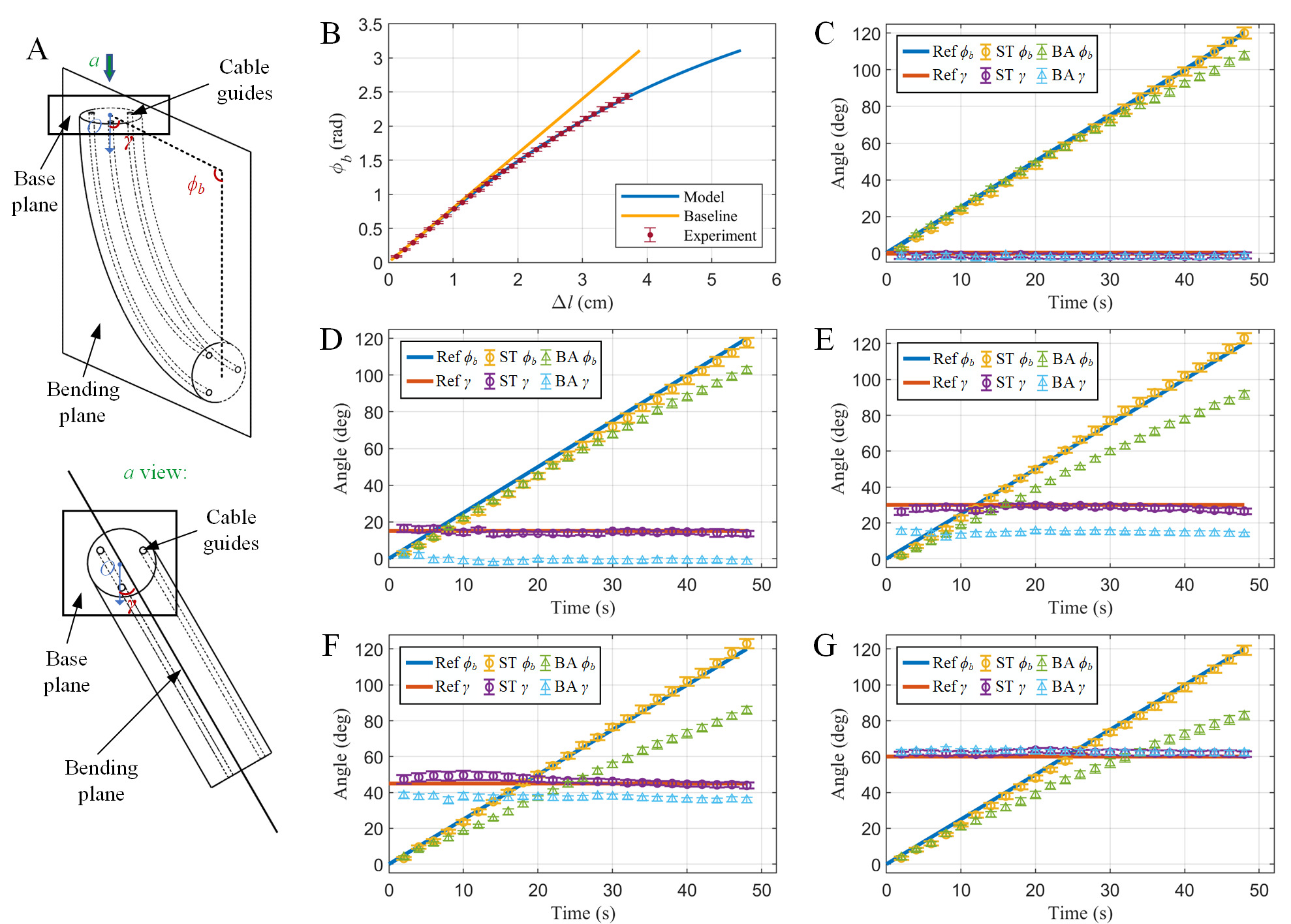}
  \caption{\small{\textbf{Experiment results for a single section of the soft robotic arm.} (A). The bending configuration of one section in experiments. (B). The relationship between cable contraction $\Delta l = L-l$ and bending angle $\phi_b$ with single cable actuation, comparing the predictions of the proposed model and the baseline model with the experimental data. (C). Tracking different bending angle $\phi_b$ when $\gamma = 0^{\circ}$ using the baseline model (BA) and the proposed model (ST). (D). Tracking different $\phi_b$ when $\gamma = 15^{\circ}$. (E). Tracking different $\phi_b$ when $\gamma = 30^{\circ}$. (F). Tracking different $\phi_b$ when $\gamma = 45^{\circ}$. (G). Tracking different $\phi_b$ when $\gamma = 60^{\circ}$. The error bars denote the standard deviations of three runs for each bending configuration in the results.}}
  \label{fig:figure}
  \vspace{-10pt}
\end{figure*}




\subsection{Experimental Results for a Single Section}

The single cable actuation results (Fig. 8B) showed that, compared with the baseline model, the proposed kinematic model was better in describing the nonlinear
relationship between the cable actuation and the bending angle for the
soft robotic arm, especially when the cable contraction length
was relatively large. 

After the model parameters were identified,
extensive experiments were conducted to compare the accuracy of the
baseline model and the proposed static model (equation (30)), where an
open loop control without feedback was used (Fig. 5D). A single section
of the robotic arm was used to track different bending angles
\(\phi_{b}\) in specific bending orientations \(\gamma\) by using
different models (Fig. 8A). The bending orientation $\gamma$ was sampled within 0 to 60 degrees becasue of the structural symmetry.

The cable actuation and control system as aforementioned was used for the multi-cable actuation experiments, where three and six motors were used for the single and the two section robotic arm, respectively. A simple and effective actuation strategy was used to deal with the multi-cable redundancy in each section and simplify the calculation of the static model: The tension of the cable $j$, that had the largest angle between the cable orientation and the bending direction, was set to be 0 in the static model calculation (made it to be "slack").
\begin{equation}
    T_j = 0, \quad j=\underset{i}{\operatorname{argmax}} \beta_i
\end{equation}

During the experiments, the target cable lengths $l_d$ were firstly calculated by using the kinematic model and the bending angle references with the help of the nonlinear system solver “$fsolve$” in MATLAB. Then, the target cable lengths $l_d$ were sent to the microcontroller and converted into the references for the rotation positions of the stepper motors. Finally, the microcontroller drove the stepper motors to reach the target rotation position and actuated the cables, where the speeds of the stepper motors were controlled to be relatively slow for a quasi-static condition. The error bars in the experimental results represent the standard deviations of the results in three identical trials. 

The experiment results of the bending configurations (\(\phi_{b}\),
\(\gamma\)) of the single section arm controlled by multiple cables by
using different models are shown in Fig. 8C-G. In the experiments, it
was shown that the tracking accuracy of the proposed static model was
significantly better compared to the baseline model in the experimental
range (Fig. 8C-G), indicating the importance of considering the
transverse deformation of the cable in the proposed robotic arm. In
particular, the tracking errors in \(\phi_{b}\) and \(\gamma\) were
small for different bending configurations when the proposed model was
used. In comparison, the tracking error in \(\phi_{b}\) for the baseline
model increased together with the target \(\phi_{b}\) when the target
\(\gamma\) was fixed, while the tracking error in \(\gamma\) for the
baseline model was almost constant despite the changing of the target
\(\phi_{b}\) when the target \(\gamma\) was fixed.

Moreover, it was also shown that in the experiment range, the tracking
error in \(\phi_{b}\) for the baseline model increased with larger
target \(\gamma\) when the target \(\phi_{b}\) was fixed. The maximum
\(\phi_{b}\) tracking error increased from about 12 degrees to about 37
degrees when the target \(\gamma\) increased from 0 to 60 degrees,
respectively. The \(\gamma\) tracking error for the baseline model
increased from about 0 to about 16 degrees when the target \(\gamma\)
increased from 0 to 30 degrees, respectively, and then decreased to near
0 when the target \(\gamma\) increased to 60 degrees. It was noticed
that for both models, the \(\gamma\) tracking error approached 0 when
target \(\gamma\) was 0 and 60 degrees, which was attributed to a single
effective cable contraction and two effective cable contractions with
the same contraction length, respectively, making the \(\gamma\)
tracking error increased and then decreased in the experimental range.
Specifically, when the target \(\gamma\) was 0 degree, there was only
one effective actuation cable (the other two cable were almost slack)
for both the baseline model and the proposed model, and the bending
orientation \(\gamma\) in the experiment naturally stayed around 0
degree, which was the same as the orientation of the actuation cable.
When the target \(\gamma\) was 60 degrees, the two actuation cables
shared the same contraction length (the third cable was almost slack)
for both the baseline model and proposed model, and \(\gamma\) in the
experiment stayed near 60 degrees because of the actuation and structure
symmetry.

A trajectory tracking experiment for the endpoint of the single section
was further conducted where the trajectory reference was included in the
workspace of the single section. The experiment showed that the average
tracking error of the single section (1.58 cm) by using the proposed
model (equation (30)) was about 35\% smaller than that (2.43 cm) of the
baseline model (Fig. 9A), and the soft section had flexible and
versatile bending configurations (Fig. 9B-C). More experiment results for the trajectory tracking experiments for the single section arm showed the repeatability of the tests and the effusiveness of the model (Fig. A1 in Appendix).  A video of these experiments and those in Section IV-E can be viewed at \url{https://youtu.be/I-e1PxHwG1Y}.

\begin{figure*}[!ht]
  \centering
\includegraphics[width=\textwidth]{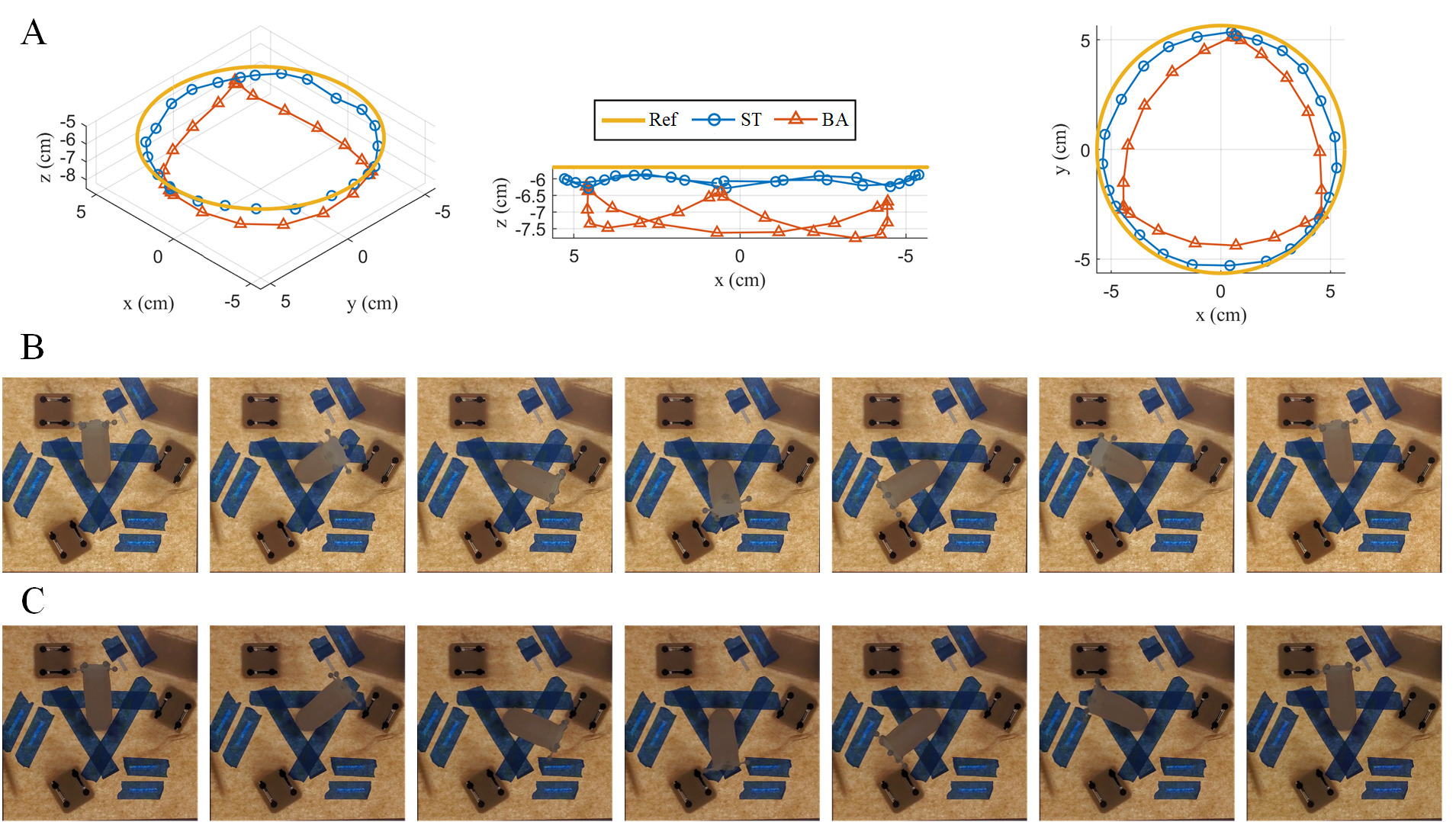}
  \caption{\small{\textbf{Experiment results for a single section of the soft robotic arm tracking a circular trajectory.} (A). Trajectories of the end position of the single section by using the baseline model (BA) and the proposed model (ST). (B). Movement and bending configurations of the single section by using BA. (C). Movement and bending configurations of the single section by using ST.}}
  \label{fig:figure}
  \vspace{-10pt}
\end{figure*}

\subsection{Experimental Results for a Two-Section Soft Robotic Arm}

A multi-section arm was assembled and utilized for the comparison of the
baseline model and the proposed model, and its performance was further
evaluated. For simplicity, a two-section arm was assembled and
controlled to track a circular trajectory within its workspace by using
the baseline model and the proposed model. The experiment results showed that, the average tracking error with the proposed model (3.76 cm) was
about 36\% smaller compared with the baseline model (5.92 cm), and the
trajectory achieved with the controller based on the proposed model was
closer to the reference (Fig. 10A-C). Compared with the tracking error
when using a single section, the tracking error of the two-section soft
arm was larger, which might be attributed to the error accumulation of
multiple sections and a more prominent gravity influence for the base
section of the soft arm.
\begin{figure*}[!ht]
  \centering
\includegraphics[width=0.95\textwidth]{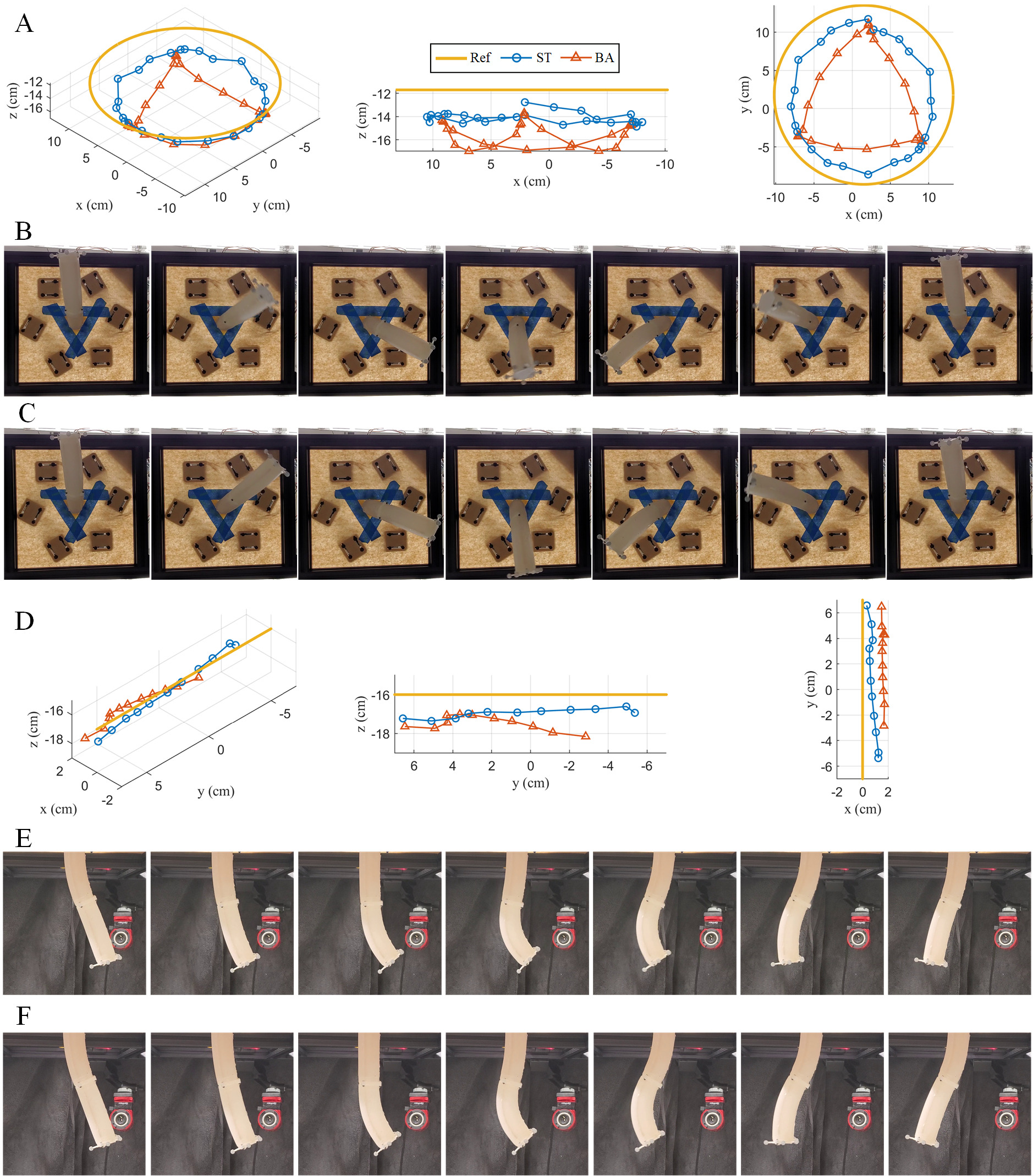}
  \caption{\small{\textbf{Experiment results for a two-section soft robotic arm.} (A). Trajectories of the end position of the arm tracking a circular path by using the baseline model (BA) and the proposed static model (ST). (B). Movement and bending configurations of the arm tracking a circular path by using BA. (C). Movement and bending configurations of the arm tracking a circular path by using ST. (D). Trajectories of the end position of the arm tracking a straight path by using BA and ST. (E). Movement and bending configurations of the arm tracking a straight path by using BA. (F). Movement and bending configurations of the arm tracking a straight path by using ST.}}
  \label{fig:figure}
  \vspace{-5pt}
\end{figure*}

In addition, the two-section arm was controlled to track a straight
trajectory within its workspace by using the two different models, where
the average tracking error (1.70 cm) of the proposed model was about
52\% smaller than that of the baseline model (3.52 cm) (Fig. 10D-F). In
summary, the extensive experiments showed the advantage of the proposed
kinematic model as compared to the geometric baseline model and
validated the flexibility and dexterity of the proposed soft robotic
arm. 

More experiment results for the trajectory tracking experiments for the two section arm showed the repeatability of the tests and the effusiveness of the model (Fig. A2, A3 in Appendix).

\section{Discussion and Conclusion}

In this work, we designed an octopus-inspired soft robotic arm and
developed a novel kinematic model to characterize its flexible
movements. The modular design of the soft arm enabled longer arm
prototypes and permitted a decoupling cable actuation system for
different sections that simplified the modeling. The hybrid fabrication
method of 3D printing and casting resulted in low-cost and easy-to-build
prototypes. An analytical static model was built to capture the
transverse deformation of the cable during actuation, which was largely
ignored in the literature.

Extensive experiments were conducted to validate the proposed model and
a geometric baseline model was used for comparison. The results of
tracking experiments for a single section of the soft arm showed an
evident advantage and smaller tracking errors for the proposed model
over the baseline model in terms of bending angle, orientation, and the
end position of the arm. Experiments with a two-section arm further
supported the efficiency of the proposed model in tracking circular and
straight trajectories for the endpoint and demonstrated the dexterity of
the proposed soft arm.

We note that our modeling approach was not only motivated by and
particularly relevant to the proposed modular soft robotic arm, but also
applicable to many other cable-driven robotic arms, especially those not
using rigid spacers, examples of which are abundant
\cite{r42,r43,r44}. Even for cable-driven soft robotic arms with multiple rigid spacers, the transverse deformation phenomenon would still exist in the sections between the rigid spacers where the cables and the soft body interact directly. 

For future work, we will extend the analytical model for the proposed soft robotic arm in several directions. First, in this work we used the PCC assumption, which would hold only in the absence of pronounced gravity effect and external forces. We plan to improve the model by considering the moments introduced by the gravity and external forces, using an iterative approach similar to the modeling of a soft pneumatic actuator by Fairchild et al.\cite{r45} In particular, one can first use the model presented in this paper to obtain the constant curvature for each section in the absence of the gravity and external forces — or equivalently, the “actuation moment” that is uniform throughout the section. Then the moments induced by the gravity and external forces can be evaluated based on that curvature, and these moments (along with the “actuation moment”) are used to update the curvature (now non-constant). This process can repeat until the curvature function converges. We will also explore the related dynamic model with external interactions. Finally, we will develop integrated embedded sensors (e.g., soft strain sensors) for the soft robotic arm, so that real-time bending configuration data for each section of the arm are made available for feedback control.

\bibliographystyle{ieeetr}
\bibliography{ref.bib}

\begin{thebibliography}{10}

\bibitem{r1}
Z.~Gong, X.~Fang, X.~Chen, {\em et~al.}, ``A soft manipulator for efficient delicate grasping in shallow water: Modeling, control, and real-world experiments,'' {\em The International Journal of Robotics Research}, vol.~40, no.~1, pp.~449--469, 2021.

\bibitem{r2}
Z.~Xie, A.~G. Domel, N.~An, {\em et~al.}, ``Octopus arm-inspired tapered soft actuators with suckers for improved grasping,'' {\em Soft robotics}, vol.~7, no.~5, pp.~639--648, 2020.

\bibitem{r3}
L.~Zongxing, L.~Wanxin, and Z.~Liping, ``Research development of soft manipulator: A review,'' {\em Advances in Mechanical Engineering}, vol.~12, no.~8, p.~1687814020950094, 2020.

\bibitem{r4}
C.~Lee, M.~Kim, Y.~J. Kim, {\em et~al.}, ``Soft robot review,'' {\em International Journal of Control, Automation and Systems}, vol.~15, pp.~3--15, 2017.

\bibitem{r5}
P.~Palmieri, M.~Melchiorre, and S.~Mauro, ``Design of a lightweight and deployable soft robotic arm,'' {\em Robotics}, vol.~11, no.~5, p.~88, 2022.

\bibitem{r6}
X.~Liang, H.~Cheong, Y.~Sun, J.~Guo, C.~K. Chui, and C.-H. Yeow, ``Design, characterization, and implementation of a two-dof fabric-based soft robotic arm,'' {\em IEEE Robotics and Automation Letters}, vol.~3, no.~3, pp.~2702--2709, 2018.

\bibitem{r7}
X.~Wang, H.~Kang, H.~Zhou, W.~Au, M.~Y. Wang, and C.~Chen, ``Development and evaluation of a robust soft robotic gripper for apple harvesting,'' {\em Computers and Electronics in Agriculture}, vol.~204, p.~107552, 2023.

\bibitem{r8}
M.~Wu, X.~Zheng, R.~Liu, N.~Hou, W.~H. Afridi, R.~H. Afridi, X.~Guo, J.~Wu, C.~Wang, and G.~Xie, ``Glowing sucker octopus (stauroteuthis syrtensis)-inspired soft robotic gripper for underwater self-adaptive grasping and sensing,'' {\em Advanced Science}, vol.~9, no.~17, p.~2104382, 2022.

\bibitem{r9}
M.~Cianchetti, T.~Ranzani, G.~Gerboni, I.~De~Falco, C.~Laschi, and A.~Menciassi, ``Stiff-flop surgical manipulator: Mechanical design and experimental characterization of the single module,'' in {\em 2013 IEEE/RSJ international conference on intelligent robots and systems}, pp.~3576--3581, IEEE, 2013.

\bibitem{r10}
A.~Diodato, M.~Brancadoro, G.~De~Rossi, {\em et~al.}, ``Soft robotic manipulator for improving dexterity in minimally invasive surgery,'' {\em Surgical innovation}, vol.~25, no.~1, pp.~69--76, 2018.

\bibitem{r11}
Z.~Wang, S.~Hirai, and S.~Kawamura, ``Challenges and opportunities in robotic food handling: A review,'' {\em Frontiers in Robotics and AI}, vol.~8, p.~789107, 2022.

\bibitem{r12}
X.~Chen, X.~Zhang, Y.~Huang, L.~Cao, and J.~Liu, ``A review of soft manipulator research, applications, and opportunities,'' {\em Journal of Field Robotics}, vol.~39, no.~3, pp.~281--311, 2022.

\bibitem{r13}
J.~Walker, T.~Zidek, C.~Harbel, S.~Yoon, F.~S. Strickland, S.~Kumar, and M.~Shin, ``Soft robotics: A review of recent developments of pneumatic soft actuators,'' in {\em Actuators}, vol.~9, p.~3, MDPI, 2020.

\bibitem{r14}
X.~Qi, H.~Shi, T.~Pinto, and X.~Tan, ``A novel pneumatic soft snake robot using traveling-wave locomotion in constrained environments,'' {\em IEEE Robotics and Automation Letters}, vol.~5, no.~2, pp.~1610--1617, 2020.

\bibitem{r15}
Z.~Xing, J.~Zhang, D.~McCoul, Y.~Cui, L.~Sun, and J.~Zhao, ``A super-lightweight and soft manipulator driven by dielectric elastomers,'' {\em Soft robotics}, vol.~7, no.~4, pp.~512--520, 2020.

\bibitem{r16}
I.~A. Anderson, T.~A. Gisby, T.~G. McKay, B.~M. O’Brien, and E.~P. Calius, ``Multi-functional dielectric elastomer artificial muscles for soft and smart machines,'' {\em Journal of applied physics}, vol.~112, no.~4, 2012.

\bibitem{r17}
H.~Yang, M.~Xu, W.~Li, and S.~Zhang, ``Design and implementation of a soft robotic arm driven by sma coils,'' {\em IEEE Transactions on Industrial Electronics}, vol.~66, no.~8, pp.~6108--6116, 2018.

\bibitem{r18}
C.~Laschi, M.~Cianchetti, B.~Mazzolai, L.~Margheri, M.~Follador, and P.~Dario, ``Soft robot arm inspired by the octopus,'' {\em Advanced robotics}, vol.~26, no.~7, pp.~709--727, 2012.

\bibitem{r19}
Y.~Kim, G.~A. Parada, S.~Liu, and X.~Zhao, ``Ferromagnetic soft continuum robots,'' {\em Science Robotics}, vol.~4, no.~33, p.~eaax7329, 2019.

\bibitem{r20}
C.~Li and C.~D. Rahn, ``Design of continuous backbone, cable-driven robots,'' {\em J. Mech. Des.}, vol.~124, no.~2, pp.~265--271, 2002.

\bibitem{r21}
W.~Dou, G.~Zhong, J.~Cao, Z.~Shi, B.~Peng, and L.~Jiang, ``Soft robotic manipulators: Designs, actuation, stiffness tuning, and sensing,'' {\em Advanced Materials Technologies}, vol.~6, no.~9, p.~2100018, 2021.

\bibitem{r22}
T.~Deng, H.~Wang, W.~Chen, X.~Wang, and R.~Pfeifer, ``Development of a new cable-driven soft robot for cardiac ablation,'' in {\em 2013 IEEE International Conference on Robotics and Biomimetics (ROBIO)}, pp.~728--733, IEEE, 2013.

\bibitem{r23}
S.~M. Mustaza, Y.~Elsayed, C.~Lekakou, C.~Saaj, and J.~Fras, ``Dynamic modeling of fiber-reinforced soft manipulator: A visco-hyperelastic material-based continuum mechanics approach,'' {\em Soft robotics}, vol.~6, no.~3, pp.~305--317, 2019.

\bibitem{r24}
Q.~Xie, T.~Wang, and S.~Zhu, ``Simplified dynamical model and experimental verification of an underwater hydraulic soft robotic arm,'' {\em Smart Materials and Structures}, vol.~31, no.~7, p.~075011, 2022.

\bibitem{r25}
R.~J. Webster~III and B.~A. Jones, ``Design and kinematic modeling of constant curvature continuum robots: A review,'' {\em The International Journal of Robotics Research}, vol.~29, no.~13, pp.~1661--1683, 2010.

\bibitem{r26}
D.~B. Camarillo, C.~F. Milne, C.~R. Carlson, M.~R. Zinn, and J.~K. Salisbury, ``Mechanics modeling of tendon-driven continuum manipulators,'' {\em IEEE transactions on robotics}, vol.~24, no.~6, pp.~1262--1273, 2008.

\bibitem{r27}
D.~B. Camarillo, C.~R. Carlson, and J.~K. Salisbury, ``Configuration tracking for continuum manipulators with coupled tendon drive,'' {\em IEEE transactions on robotics}, vol.~25, no.~4, pp.~798--808, 2009.

\bibitem{r28}
F.~Renda, M.~Giorelli, M.~Calisti, M.~Cianchetti, and C.~Laschi, ``Dynamic model of a multibending soft robot arm driven by cables,'' {\em IEEE Transactions on Robotics}, vol.~30, no.~5, pp.~1109--1122, 2014.

\bibitem{r29}
T.~Morales~Bieze, A.~Kruszewski, B.~Carrez, and C.~Duriez, ``Design, implementation, and control of a deformable manipulator robot based on a compliant spine,'' {\em The International Journal of Robotics Research}, vol.~39, no.~14, pp.~1604--1619, 2020.

\bibitem{r30}
S.~Grazioso, G.~Di~Gironimo, and B.~Siciliano, ``A geometrically exact model for soft continuum robots: The finite element deformation space formulation,'' {\em Soft robotics}, vol.~6, no.~6, pp.~790--811, 2019.

\bibitem{r31}
D.~Trivedi, A.~Lotfi, and C.~D. Rahn, ``Geometrically exact models for soft robotic manipulators,'' {\em IEEE Transactions on Robotics}, vol.~24, no.~4, pp.~773--780, 2008.

\bibitem{r32}
F.~Xu, H.~Wang, K.~W.~S. Au, W.~Chen, and Y.~Miao, ``Underwater dynamic modeling for a cable-driven soft robot arm,'' {\em IEEE/ASME transactions on Mechatronics}, vol.~23, no.~6, pp.~2726--2738, 2018.

\bibitem{r33}
D.~C. Rucker and R.~J. Webster~III, ``Statics and dynamics of continuum robots with general tendon routing and external loading,'' {\em IEEE Transactions on Robotics}, vol.~27, no.~6, pp.~1033--1044, 2011.

\bibitem{r34}
F.~Renda, C.~Armanini, V.~Lebastard, F.~Candelier, and F.~Boyer, ``A geometric variable-strain approach for static modeling of soft manipulators with tendon and fluidic actuation,'' {\em IEEE Robotics and Automation Letters}, vol.~5, no.~3, pp.~4006--4013, 2020.

\bibitem{r35}
F.~Renda, C.~Armanini, A.~Mathew, and F.~Boyer, ``Geometrically-exact inverse kinematic control of soft manipulators with general threadlike actuators’ routing,'' {\em IEEE Robotics and Automation Letters}, vol.~7, no.~3, pp.~7311--7318, 2022.

\bibitem{r36}
F.~Renda, F.~Boyer, J.~Dias, and L.~Seneviratne, ``Discrete cosserat approach for multisection soft manipulator dynamics,'' {\em IEEE Transactions on Robotics}, vol.~34, no.~6, pp.~1518--1533, 2018.

\bibitem{r37}
Y.~Wu, J.~K. Yim, J.~Liang, Z.~Shao, M.~Qi, J.~Zhong, Z.~Luo, X.~Yan, M.~Zhang, X.~Wang, {\em et~al.}, ``Insect-scale fast moving and ultrarobust soft robot,'' {\em Science robotics}, vol.~4, no.~32, p.~eaax1594, 2019.

\bibitem{r38}
H.~R. Choi, K.~Jung, S.~Ryew, J.-D. Nam, J.~Jeon, J.~C. Koo, and K.~Tanie, ``Biomimetic soft actuator: design, modeling, control, and applications,'' {\em IEEE/ASME transactions on mechatronics}, vol.~10, no.~5, pp.~581--593, 2005.

\bibitem{r39}
X.~Qi, T.~Gao, and X.~Tan, ``Bioinspired 3d-printed snakeskins enable effective serpentine locomotion of a soft robotic snake,'' {\em Soft Robotics}, vol.~10, no.~3, pp.~568--579, 2023.

\bibitem{r40}
S.~Kim, C.~Laschi, and B.~Trimmer, ``Soft robotics: a bioinspired evolution in robotics,'' {\em Trends in biotechnology}, vol.~31, no.~5, pp.~287--294, 2013.

\bibitem{r41}
L.~Sciavicco and B.~Siciliano, {\em Modelling and control of robot manipulators}.
\newblock Springer Science \& Business Media, 2001.

\bibitem{r42}
H.~Wang, W.~Chen, X.~Yu, T.~Deng, X.~Wang, and R.~Pfeifer, ``Visual servo control of cable-driven soft robotic manipulator,'' in {\em 2013 IEEE/RSJ International Conference on Intelligent Robots and Systems}, pp.~57--62, IEEE, 2013.

\bibitem{r43}
T.~Deng, H.~Wang, W.~Chen, X.~Wang, and R.~Pfeifer, ``Development of a new cable-driven soft robot for cardiac ablation,'' in {\em 2013 IEEE International Conference on Robotics and Biomimetics (ROBIO)}, pp.~728--733, IEEE, 2013.

\bibitem{r44}
W.~Dou, G.~Zhong, J.~Cao, Z.~Shi, B.~Peng, and L.~Jiang, ``Soft robotic manipulators: Designs, actuation, stiffness tuning, and sensing,'' {\em Advanced Materials Technologies}, vol.~6, no.~9, p.~2100018, 2021.

\bibitem{r45}
P.~Fairchild, N.~Shepard, Y.~Mei, and X.~Tan, ``Semi-physical modeling of soft pneumatic actuators with stiffness tuning,'' {\em ASME Letters in Dynamic Systems and Control}, pp.~1--6, 2023.

\end{thebibliography}

\newpage
\renewcommand\thefigure{A\arabic{figure}} 
\setcounter{figure}{0} 
\appendix
\subsection{Supplementary Derivations for Modeling of Soft Robotic Arm}
The detailed derivation for the equation (14) is as follows:
We can first denote $\alpha = \phi_b /2$ and $\psi = \phi_b - \theta_0$, then we can derive:
\begin{equation*}
\begin{aligned}
\boldsymbol{M}&_{e q} +\boldsymbol{M}_T\\
= & r_{e q} \times F_{e q}+r_T \times F_T \\
= & {\begin{bmatrix}
-d-R_d \sin ^2\left(\phi_b / 2\right) \\
R_d \sin \left(\phi_b / 2\right) \cos \left(\phi_b / 2\right)
\end{bmatrix} \times T\begin{bmatrix}
-\sin \left(\phi_b-\theta_0\right)+\sin \theta_0 \\
\cos \left(\phi_b-\theta_0\right)-\cos \theta_0
\end{bmatrix} } \\
& +\begin{bmatrix}{c}
-d-2 R_d \sin ^2\left(\phi_b / 2\right) \\
2 R_d \sin \left(\phi_b / 2\right) \cos \left(\phi_b / 2\right)
\end{bmatrix} \times T\begin{bmatrix}{c}
\sin \left(\phi_b-\theta_0\right) \\
-\cos \left(\phi_b-\theta_0\right)
\end{bmatrix} \\
= & T \cdot(d \cos \theta_0-d \cos \psi-R_d \sin ^2 \alpha \cos \psi + R_d \sin ^2 \alpha \cos \theta_0 \\
& +R_d \sin \alpha \cos \alpha \sin \psi 
-R_d \sin \alpha \cos \alpha \sin \theta_0+d \cos \psi\\
& +2 R_d \sin ^2 \alpha \cos \psi-2 R_d \sin \alpha \cos \alpha \sin \psi) \\
= & T \cdot(d \cos \theta_0+R_d \sin ^2 \alpha \cos \psi 
+R_d \sin ^2 \alpha \cos \theta_0\\
& -R_d \sin \alpha \cos \alpha \sin \theta_0-R_d \sin \alpha \cos \alpha \sin \psi) \\
= & T \cdot\left(d \cos _0+R_d \cdot D\right)
\end{aligned}
\end{equation*}
where $D$ is calculated as:
\begin{equation*}
\begin{aligned}
D=&\sin ^2 \alpha \cos \theta_0+\sin ^2 \alpha \cos \psi-\sin \alpha \cos \alpha \sin \theta_0 \\
&-\sin \alpha \cos \alpha \sin \psi \\
=&\frac{1-\cos \phi_b}{2} \cos \theta_0+\frac{1-\cos \phi_b}{2} \cos \phi_b \cos \theta_0 \\
& +\frac{1-\cos \phi_b}{2} \sin \phi_b \sin \theta_0 
-\frac{\sin \phi_b}{2} \sin \theta_0 \\
& -\frac{\sin \phi_b \sin \phi_b}{2} \cos \theta_0+\frac{\sin \phi_b \cos \phi_b}{2} \sin \theta_0 \\
=&\frac{1}{2} \cos \theta_0-\frac{\cos \phi_b \cos \theta_0}{2}+\frac{\cos \phi_b \cos \theta_0}{2}\\
&-\frac{\cos \phi_b \cos \phi_b \cos \theta_0}{2}+\frac{\sin \phi_b \sin \theta_0}{2} \\
& -\frac{\cos \phi_b \sin \phi_b \sin \theta_0}{2}-\frac{\sin \phi_b \sin \theta_0}{2}\\
&-\frac{\sin \phi_b \sin \phi_b \cos \theta_0}{2} +\frac{\cos \phi_b \sin \phi_b \sin \theta_0}{2} \\
=&\frac{1}{2} \cos \theta_0-\frac{\cos ^2 \phi_b \cos \theta_0}{2}-\frac{\sin ^2 \phi_b \cos \theta_0}{2} \\
=&0 \\
\end{aligned}
\end{equation*}

Thus, finally we can derive:
\begin{equation*}
\boldsymbol{M}_{e q}+\boldsymbol{M}_T=T d \cos \theta_0
\end{equation*}
which shows an elegant result of the external moment applied to the single section of the cable driven robotic arm.

\subsection{Supplementary Results for Trajectory Tracking Experiments}
More trajectory tracking experiments were conducted for the proposed soft robotic (Fig. A1-A3). Two more sets of experiments were conducted for a single section of the soft arm tracking the circular trajectory (Fig. A1A-B), which shared the same experiment settings as the experiments in Fig. 9.
The result of the supplementary experiment set 1 showed that the average tracking error of the single section by using the proposed model (1.55 cm) was about 49$\%$ smaller than that of the baseline model (3.06 cm) (Fig. A1A). The result of the supplementary experiment set 2 showed that the average tracking error of the single section by using the proposed model (1.57 cm) was about 43$\%$ smaller than that of the baseline model (2.75 cm) (Fig. A1B). The supplementary experiment sets demonstrated similar results as the trajectory tracking results of the experiments in Fig. 9 and showed their reproducibility.

\begin{figure*}[!ht]
  \centering
\includegraphics[width=0.75\textwidth]{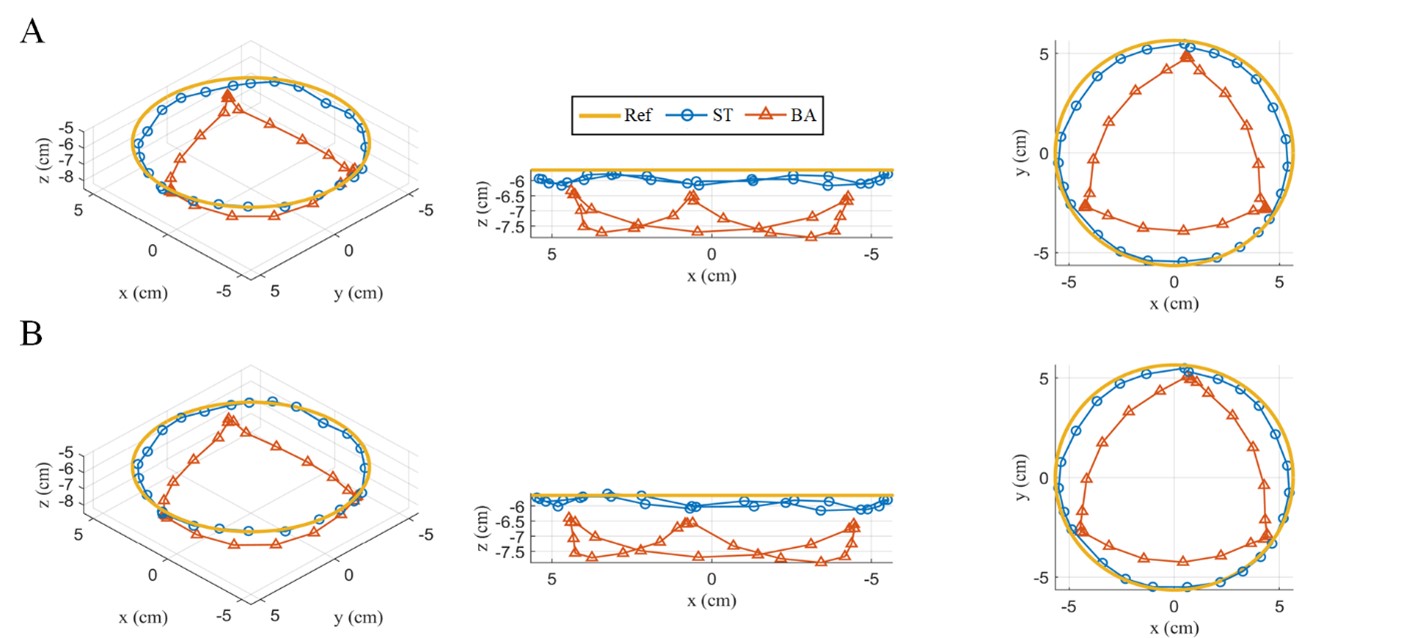}
  \caption{\small{\textbf{More experiment results for a single section of the soft robotic arm tracking a circular trajectory. }(A). Trajectories of the end position of the single section by using the baseline model (BA) and the proposed model (ST) in the supplementary experiment set 1. (B). Trajectories of the end position of the single section by using the baseline model (BA) and the proposed model (ST) in the supplementary experiment set 2.}}
  \label{fig:figure}
  \vspace{-10pt}
\end{figure*}

Furthermore, two more sets of experiments were conducted for the two-section soft robotic arm tracking the circular trajectory (Fig. A2A-B) and the straight trajectory (Fig. A3A-B), where the experiment settings remained the same as those in the main text. 
In the experiments tracking the circular trajectory, the average tracking error with the proposed model (3.88 cm) was about 36$\%$ smaller compared with the baseline model (6.02 cm) (Fig. A2A) in the supplementary experiment set 1. For the supplementary experiment set 2, the average tracking error with the proposed model (3.84 cm) was about 35$\%$ smaller compared with the baseline model (5.93 cm) (Fig. A2B). The supplementary experiment sets demonstrated similar results as the trajectory tracking results of the experiments in Fig. 10A and showed their reproducibility.

\begin{figure*}[!ht]
  \centering
\includegraphics[width=0.75\textwidth]{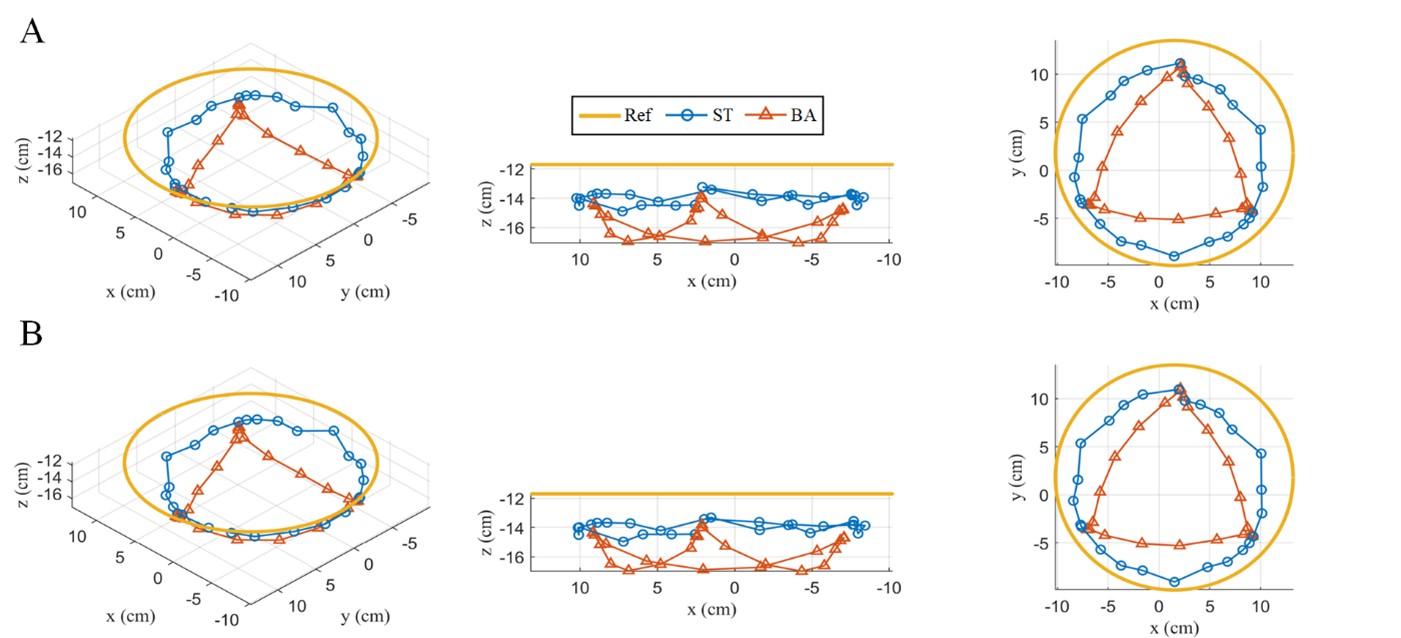}
  \caption{\small{\textbf{More experiment results for a two-section soft robotic arm tracking a circular trajectory.}(A). Trajectories of the end position of the two-section arm by using the baseline model (BA) and the proposed model (ST) in the supplementary experiment set 1. (B). Trajectories of the end position of the two-section arm by using the baseline model (BA) and the proposed model (ST) in the supplementary experiment set 2.}}
  \label{fig:figure}
  \vspace{-10pt}
\end{figure*}

In the experiments tracking the straight trajectory, the average tracking error with the proposed model (1.77 cm) was about 41$\%$ smaller compared with the baseline model (2.98 cm) (Fig. A3A) in the supplementary experiment set 1. For the supplementary experiment set 2, the average tracking error with the proposed model (1.57 cm) was about 47$\%$ smaller compared with the baseline model (2.96 cm) (Fig. A3B). The supplementary experiment sets demonstrated similar results as the trajectory tracking results of the experiments in Fig. 10D and showed their reproducibility.

\begin{figure*}[!ht]
  \centering
\includegraphics[width=0.75\textwidth]{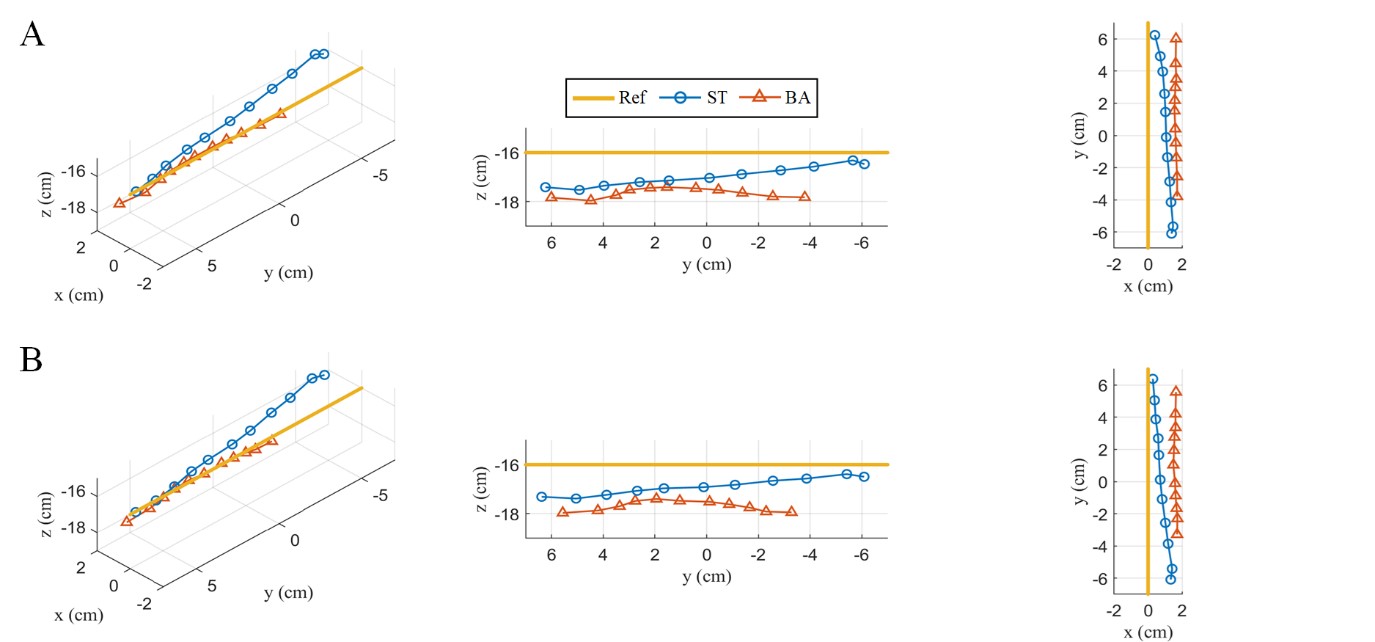}
  \caption{\small{\textbf{More experiment results for a two-section soft robotic arm tracking a straight trajectory.}(A). Trajectories of the end position of the two-section arm by using the baseline model (BA) and the proposed model (ST) in the supplementary experiment set 1. (B). Trajectories of the end position of the two-section arm by using the baseline model (BA) and the proposed model (ST) in the supplementary experiment set 2.}}
  \label{fig:figure}
  \vspace{-10pt}
\end{figure*}

\end{document}